\documentclass[times,twocolumn,review]{elsarticle}

\usepackage{medima}
\usepackage{framed,multirow}

\usepackage{amssymb}
\usepackage{latexsym}

\usepackage{url}
\usepackage{xcolor}

\usepackage{hyperref}
\usepackage{bbm}
\usepackage{amsmath,amsfonts,bm}
\def\rvx{{\mathbf{x}}}
\def\rvz{{\mathbf{z}}}
\def\rvy{{\mathbf{y}}}
\def\rvc{{\mathbf{c}}}
\def\rvv{{\mathbf{v}}}
\newcommand\etal {{\it et al.}}
\newcommand{\KL}{D_{\mathrm{KL}}}
\newcommand{\E}{\mathbb{E}}

\def\gL{{\mathcal{L}}}
\newcommand\ie {{\it i.e., }}

\newcommand\eg {{\it e.g., }}

\newcommand{\blue}[1]{\textcolor[RGB]{0, 0, 0}{#1}}

\usepackage{stackengine}
\def\delequal{\mathrel{\ensurestackMath{\stackon[1pt]{=}{\scriptstyle\Delta}}}}

\definecolor{newcolor}{rgb}{.8,.349,.1}

\journal{Medical Image Analysis}

\begin{document}

\verso{Sumedha Singla \textit{et~al.}}

\begin{frontmatter}

\title{Explaining the Black-box Smoothly -\
A Counterfactual Approach}%

\author[1]{Sumedha \snm{Singla}\corref{cor1}}
\cortext[cor1]{Corresponding author: 
 Email: sumedha.singla@pitt.edu;}
 \author[2]{Motahhare \snm{Eslami}}
\ead{meslami@andrew.cmu.edu}
\author[3]{Brian  \snm{Pollack}\fnref{fn1}}
\ead{brp98@pitt.edu}
\fntext[fn1]{This work was done when author was in University of Pittsburgh.}
\author[4]{Stephen \snm{Wallace}}
\ead{wallacesr2@upmc.edu}
\author[3]{Kayhan \snm{Batmanghelich}}
\ead{kayhan@pitt.edu}
\address[1]{Computer Science Department at the University of Pittsburgh, Pittsburgh, PA, 15206, USA}
\address[2]{School of Computer Science, 
Human-Computer Interaction Institute, 
Carnegie Mellon University}
\address[3]{Department of Biomedical Informatics, the University of Pittsburgh, Pittsburgh, PA, 15206, USA}
\address[4]{University of Pittsburgh Medical School, Pittsburgh, PA, 15206, USA}
\received{10 21 2021}
\finalform{00 00 2021}
\accepted{00 00 2021}
\availableonline{00 00 2021}
\communicated{S. Singla}

\begin{abstract}
We propose a BlackBox \emph{Counterfactual Explainer}, designed to explain image classification models for medical applications. Classical approaches (\eg, saliency maps) that assess feature importance do not explain \emph{how} imaging features in important anatomical regions are relevant to the classification decision. Such reasoning is crucial for transparent decision-making in healthcare applications. Our framework explains the decision for a target class by gradually \emph{exaggerating} the semantic effect of the class in a query image. We adopted a Generative Adversarial Network (GAN) to generate a progressive set of perturbations to a query image, such that the classification decision changes from its original class to its negation. Our proposed loss function preserves essential details (\eg support devices) in the generated images.

We used counterfactual explanations from our framework to audit a classifier trained on a chest x-ray dataset with multiple labels. Clinical evaluation of model explanations is a challenging task. 
We proposed clinically-relevant quantitative metrics such as cardiothoracic ratio and the score of a  healthy costophrenic recess to evaluate our explanations. We used these metrics to quantify the counterfactual changes between the populations with negative and positive decisions for a diagnosis by the given classifier.

\blue{We conducted a human-grounded experiment with diagnostic radiology residents to compare different styles of explanations (no explanation, saliency map, cycleGAN explanation, and our counterfactual explanation) by evaluating different aspects of explanations: (1) understandability,  (2) classifier's decision justification, (3) visual quality, (d) identity preservation, and (5) overall helpfulness of an explanation to the users. Our results show that our counterfactual explanation  was the only explanation method that significantly improved  the users' understanding of the classifier's decision compared to the no-explanation baseline.} Our metrics established a benchmark for evaluating model explanation methods in medical images. Our explanations revealed that the classifier relied on clinically relevant radiographic features for its diagnostic decisions, thus making its decision-making process more transparent to the end-user.

\end{abstract}


\end{frontmatter}


\section{Introduction}
\label{sec:introduction}

Machine learning, specifically Deep Learning (DL), is being increasingly used for sensitive applications such as Computer-Aided Diagnosis~\citep{Hosny2018ArtificialRadiology} and other tasks in the medical imaging domain~\citep{Rajpurkar2018DeepRadiologists,AloneRadiologists}. However, for real-world deployment~\citep{Wang2020ShouldMedicine}, the decision-making process of these models should be explainable to humans to obtain their trust in the model ~\citep{Gastounioti2020IsAI,Jiang2018ToClassifier}. Explainability is essential for auditing the model~\citep{Winkler2019AssociationRecognition}, identifying various failure modes~\citep{Oakden-Rayner2020HiddenImaging,Eaton-Rosen2018TowardsPredictions} or hidden biases in the data or the model~\citep{Larrazabal2020GenderDiagnosis}, and for obtaining new insights from large-scale studies~\citep{Rubin2018LargeNetworks}.

\blue{With the advancement of DL methods for medical imaging analysis, deep neural networks (DNNs)  have achieved near-radiologist performance in multiple image classification tasks~\citep{seah2021effect,Rajpurkar2017CheXNet:Learning}. However, DNNs are criticized for their ``black-box" nature, \ie they fail to provide a simple explanation as to why a given input image produces a corresponding output~\citep{Tonekaboni2019WhatUse}.To address this concern, multiple model explanation techniques have been proposed that aim to explain the decision-making process of DNNs~\citep{Selvaraju2017Grad-cam:Localization,cohen2021gifsplanation}. The most common form of explanation in medical imaging is a class-specific heatmap overlaid on the input image. It highlights the most relevant regions (\emph{where}) for the classification decision~\citep{Rajpurkar2017CheXNet:Learning,Young2019DeepDermatologist}.} However, the location information alone is insufficient for applications in medical imaging. Different diagnoses may affect the same anatomical regions, resulting in similar explanations for multiple diagnosis, resulting in inconclusive explanations. A thorough explanation should explain \emph{what} imaging features are present in those important locations, and \emph{how} changing such features modifies the classification decision.

\begin{figure}
\includegraphics[width=0.95\linewidth]{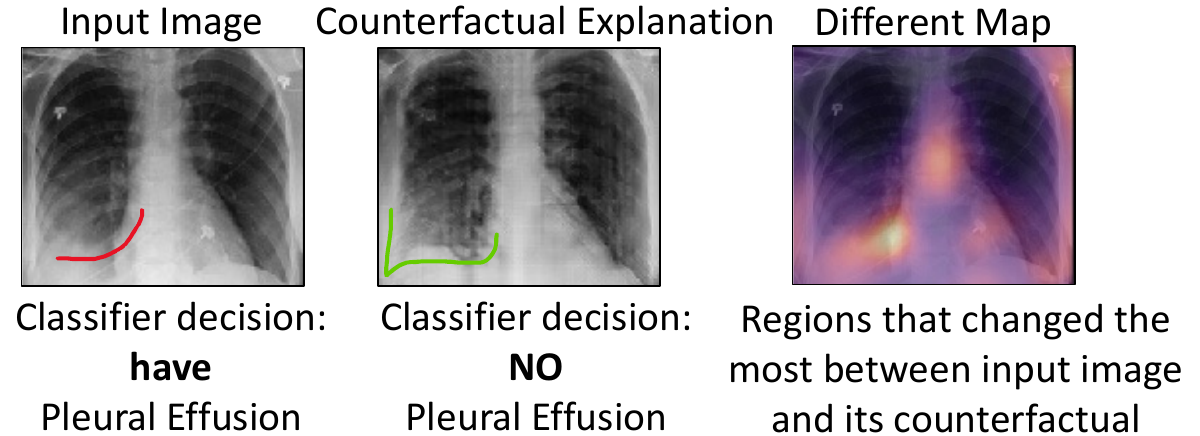}
\caption{ Counterfactual explanation shows ``where" + ``what" minimum change must be made to the input to flip the classification decision.  For Pleural Effusion, we can observe vanishing of the meniscus (red) in counterfactual image as compared to the query image.}
\label{Fig_PE}
\end{figure}

\blue{To address this problem, we propose a novel explanation method to provide a counterfactual explanation.} A counterfactual explanation is a perturbation of the input image such that the classification decision is flipped. \blue{By comparing, the input image and its corresponding counterfactual image, the end-users can visualize the difference in important image features that leads to a change in classification decision. Fig.~\ref{Fig_PE} shows an example. The input image is predicted as positive for pleural effusion (PE), while the generated counterfactual image is negative for PE. The changes are mostly concentrated in the lower lobe region, which is known to be clinically important for PE~\citep{PleuralTomography}. Counterfactual explanation is used to derive a pseudo-heat-map, highlighting the regions that change the most in the transformation (difference map in Fig.~\ref{Fig_PE}). } 

We demonstrate the performance of the counterfactual explainer on a chest x-ray (CXR) dataset. \blue{Rather than generating just one counterfactual image at the end of the prediction spectrum, our explanation function generates a series of perturbed images that gradually traverse the decision boundary from one extreme (negative decision) to another (positive decision) for a given target class.} We adopted a conditional Generative Adversarial Network (cGAN) as our explanation function~\citep{Singla2020ExplanationExaggeration}. We extend the cGAN to preserve small or uncommon details during image generation~\citep{Bau2019SeeingGenerate}. Preserving such details is particularly important in our application, as the missing information may include support devices that may influence human users' perceptions. To this end, we incorporated semantic segmentation and object detection into our loss function to preserve the shape of the anatomy and foreign objects during image reconstruction. We evaluated the quality of our explanations using different quantitative metrics, including clinical measures. Further, we performed a clinical study with 12 radiology residents to compare the explanations for the proposed method and the baseline models.

\subsection{Related work}
Posthoc \emph{explanation} is a popular approach that aims to improve human understanding of a pre-trained classifier.  Our work broadly relates to the following posthoc methods:

\emph{Feature Attribution} methods provide explanation by producing a saliency map that shows the importance of each input component (\eg pixel) to the classification decision.  \emph{Gradient-based} approaches~\citep{Simonyan2013DeepMaps,Springenberg2015StrivingNet,Bach2015OnPropagation,Shrikumar2017LearningDifferences,Sundararajan2017AxiomaticNetworks,LundbergAPredictions,Selvaraju2017Grad-cam:Localization} produce a saliency map by computing the gradient of the classifier's output with respect to the input components.  Such methods are often applied to the medical imaging studies, \eg CXR~\citep{Rajpurkar2017CheXNet:Learning}, skin imaging~\citep{Young2019DeepDermatologist}, brain MRI~\citep{Eitel2019TestingClassification} and retinopathy~\citep{Sayres2019UsingRetinopathy}.

\emph{Perturbation-based} methods identify salient regions by directly manipulating the input image and analyzing the resulting changes in the classifier's output. Such methods modify specific pixels or regions in an input image, either by masking with constant values~\citep{Dabkowski2017RealClassifiers} or with random noise, occluding~\citep{Zhou2014ObjectCnns},  localized blurring~\citep{Fong2017InterpretablePerturbation}, or in-filling~\citep{Chang2019ExplainingGeneration}.   Especially for medical images, such perturbations may introduce anatomically implausible features or textures.
Our proposed method also generates a perturbation of the query image such that classification decision is flipped. But in contrast to the above methods, we enforce consistency between the perturbed data and the real data distribution to
ensure that the perturbations are plausible and visually similar to the input.

\emph{Counterfactual Explanations} are a type of contrastive~\citep{Dhurandhar2018ExplanationsNegatives} explanation that provides a useful way to audit the classifier and determine causal attributes that lead to the classification decision~\citep{ParafitaMartinez2019ExplainingAttribution,singla2021using}. Similar to our method, generative models like GANs and variational autoencoders (VAE) are used to compute interventions that generate realistic counterfactual explanations~\citep{ cohen2021gifsplanation,Joshi2019TowardsSystems}. Much of this work is limited to simpler image datasets like MNIST, celebA~\citep{Liu2019GenerativeLearning,VanLooveren2019InterpretablePrototypes} or simulated data~\citep{ParafitaMartinez2019ExplainingAttribution}. For more complex natural images, previous studies~\citep{Chang2019ExplainingGeneration,Agarwal2019ExplainingModels} focused on finding and in-filling salient regions to generate counterfactual images. In contrast, our explanation function doesn't require any re-training for generating explanations for a new image at inference time. In another line of work~\citep{Wang2020SCOUT:Explanations,Goyal2019CounterfactualExplanations} provide counterfactual explanations that explain both the predicted and the counter class. Further, researcher  ~\citep{Narayanaswamy2020ScientificTranslation,DeGrave2020AISignal} has used a cycle-GAN~\citep{Zhu2017UnpairedNetworks} model to perform image-to-image translation between normal and abnormal images. Such methods are independent of the classifier.  In contrast, our framework uses a classifier consistency loss to enable image perturbation that is consistent with the classifier.

\subsection{Contributions}
In this paper, we propose a progressive counterfactual explainer, that explains the decision of a pre-trained image classifier. Our contributions are summarized as follows:
\begin{enumerate}
    \item \blue{We developed a cGAN-based framework to generate progressively changing perturbations of the query image, such that classification decision changes from being negative to being positive for a given target class.}
   \item \blue{Our method preserved the anatomical shape and foreign  objects such as support devices across generated images by adding a specialized reconstruction loss. The loss incorporates  context from semantic segmentation and foreign object detection networks.} 
    \item We performed a thorough qualitative and quantitative evaluation of our explanation function to audit a classifier trained on a CXR dataset.
      \item We proposed quantitative metrics based on clinical definition of two diseases (cardiomegaly and PE). We are one of the first methods to use such metrics for quantifying DNN model explanation. Specifically, we used these metrics to quantify statistical differences between the real images and their corresponding counterfactual images.
      
      \blue{\item We are one of the first methods to conduct a thorough human-grounded study to evaluate different counterfactual explanations for medical imaging task. Specifically, we collected and compared feedback from diagnostic radiology residents, on different aspects of explanations: (1) understandability,  (2) classifier's decision justification, (3) visual quality, (d) identity preservation, and (5) overall helpfulness of an explanation to the users.}
\end{enumerate}

\section{Methodology}
\blue{We consider a \emph{black-box} image classifier $f$, with high prediction accuracy.} We assume that $f$ is a differentiable function and we have access to its value as well as its gradient with respect to the input $\nabla_{\rvx} f(\rvx)$.   \blue{We also assume access to either the training data for $f$, or an equivalent dataset with competitive prediction accuracy.} 

\blue{\emph{Notation:} The classification function is defined as $f: \mathbb{R}^d \rightarrow \mathbb{R}^K$, where $d$ is the dimensionality of the image space and  $K$ is the number of classes. The classifier produces point estimates for posterior probability of class $k$ as $\mathbb{P} (y_k|\rvx) = f(\rvx)[k] \in [0,1]$. }

\blue{\emph{Explanation function: }We aim to explain the decision of function $f$ for a target class $k$. We consider visual explanation of the black-box as a generative process that produces a plausible and realistic perturbation of the query image $\rvx$ such that the classification decision for class $k$ is changed to a desired value $\mathbf{c}$. This idea allows us to view $\rvc$ as a ``knob''. By gradually changing the desired output $\mathbf{c}$ in range $[0,1]$, we generate progressively changing perturbations of 
 the query image $\rvx$, such that classification decision changes from being negative to being positive for a class $k$.}

\begin{figure}[!ht]
\centering
\includegraphics[width=1.0\linewidth]{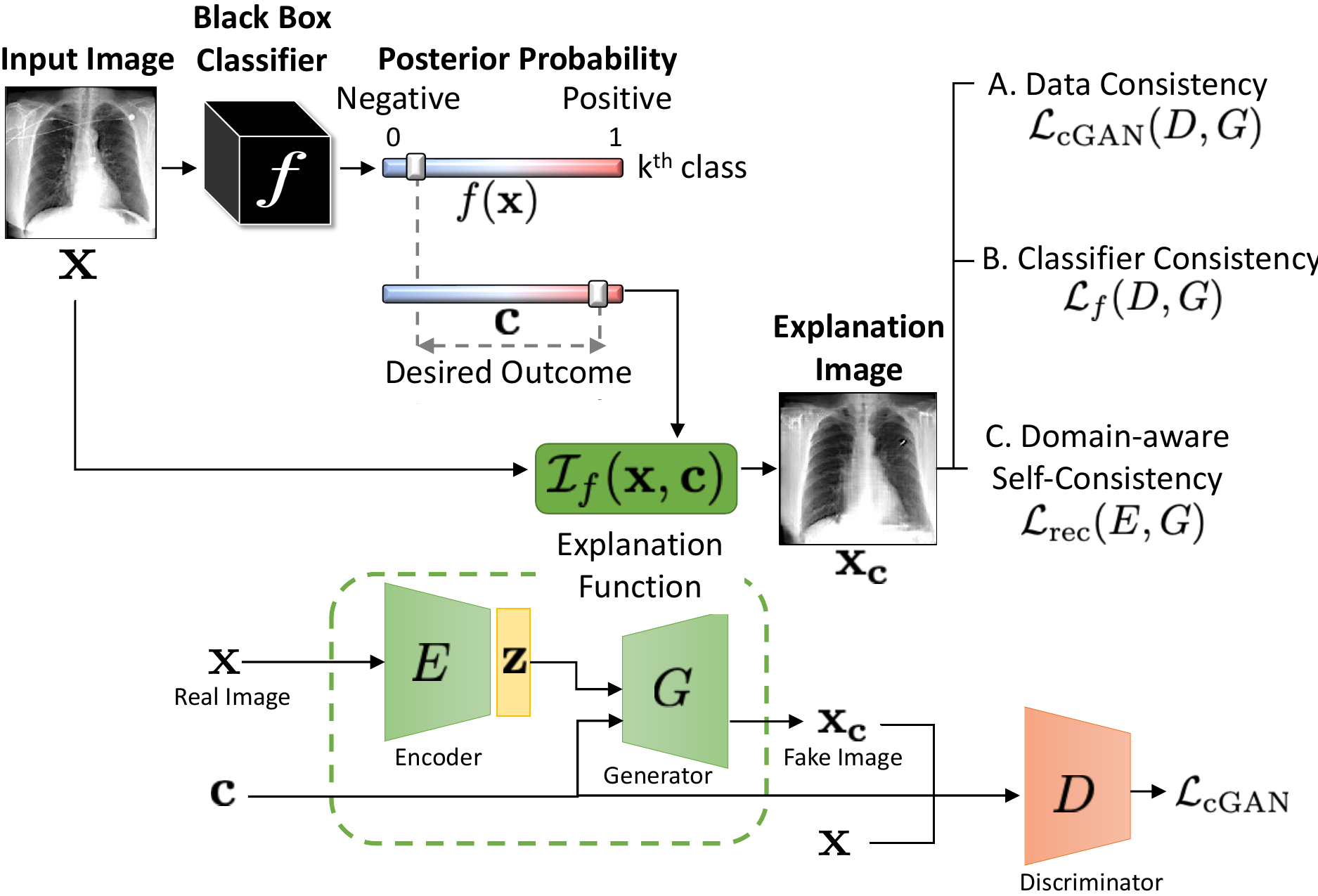}
\caption{Explanation function $\mathcal{I}_f(\rvx, \rvc)$ for classifier $f$. Given an input image $\rvx$, we  generates a perturbation of the input, $\rvx_{\rvc}$ as explanation, such that the posterior probability, $f$, changes from its original value, $f(\rvx)$, to a desired value $\rvc$ while satisfying the three consistency constraints.}
\label{Fig_Model}
\end{figure}

\blue{To achieve this, we propose  an \emph{explanation} function $\rvx_{\rvc}  \delequal \mathcal{I}_{f_k}(\rvx,\rvc): (\mathcal{X}, \mathbb{R}) \rightarrow \mathcal{X}$. This function takes two arguments: a query image $\rvx$ and the desired posterior probability $\rvc$ for the target class $k$. The explanation function generates a perturbed image $\rvx_{\rvc}$ such that $f(\rvx_{\rvc})[k] \approx \rvc$.  For simplicity, we will drop $k$ from subsequent notations.} Fig.~\ref{Fig_Model} summarizes our framework. We design the explanation function to satisfy the following properties:

(A) \textbf{Data consistency}: $\rvx_{\rvc}$ should resemble data instance from input space \ie if input space comprises of CXRs, $\rvx_{\rvc}$ should look like a CXR with minimum artifacts or blurring.

(B) \textbf{Classifier consistency}: $\rvx_{\rvc}$ should produce the desired output from the classifier $f$, \ie $f(\mathcal{I}_f(\rvx, \rvc)) \approx  \rvc$.

(C) \textbf{Context-aware self-consistency}: On using the original decision as the condition, \ie $\rvc = f(\rvx)$, the explanation function should reconstruct the query image. We forced this condition for self-consistency as $\mathcal{I}_f(\rvx, f(\rvx)) = \rvx$ and  for cyclic-consistency as $\mathcal{I}_f(\rvx_{\rvc}, f(\rvx)) = \rvx$. Further, we constrained the explanation function to achieve the aforementioned reconstructions while preserving anatomical shape and  foreign objects (\eg pacemaker) in the input image.

\emph{Overall objective: }\blue{Our explanation function $\mathcal{I}_f(\rvx, \rvc)$ is trained end to end to learn parameters for three networks, an image encoder $E(\cdot)$, a conditional GAN generator $G(\cdot)$ and a discriminator $D(\cdot)$, to satisfy the above three properties while minimizing the following objective function:}

\begin{equation}
    \min_{E, G} \max_{D} \lambda_{cGAN}  \mathcal{L}_{\text{cGAN}}(D,G) + \lambda_{f} \mathcal{L}_{f}(D,G) + \lambda_{rec}\mathcal{L}_{\text{rec}}(E,G)
    \label{eq_final}
\end{equation}

where $\mathcal{L}_{\text{cGAN}}$ is a conditional GAN-based loss function that enforces data-consistency, $\mathcal{L}_{f}$ enforces classifier consistency through a KullbackLeibler (KL) divergence loss and $\mathcal{L}_{\text{rec}}$ is a reconstruction loss that enforces self-consistency. Hyper-paramerters,  $\lambda_{cGAN}, \lambda_{f}$ and $\lambda_{rec}$ controls the balance between the terms. In the following sections, we will discuss each property and the associated loss term in detail.

\subsection{Data consistency}
We formulated the explanation function, $\mathcal{I}_f(\rvx, \rvc)$,  as an image encoder $E(\cdot)$ followed by a conditional GAN (cGAN)~\citep{Miyato2018CGANsDiscriminator}, with $\rvc$ as the condition. The encoder enables the transformation of a given image, while the GAN framework generates realistic-looking transformations as an explanation image. The cGAN is a variant of GAN that allows the conditional generation of the data by incorporating extra information as the context.
Like GANs, cGAN is composed of two deep networks, generator $G(\cdot)$  and discriminator $D(\cdot)$. 
The $G$, $D$ are trained adversarially by optimizing the following objective function,
\begin{equation}
\begin{split}
\mathcal{L}_{\text{cGAN}}(D,G)  & =  \E_{\rvx,\rvc \sim P(\rvx,\rvc)}\big[\log \big(D(\rvx, \rvc)\big)\big] +  \\
& \E_{ \rvz \sim P_{\rvz}, \rvc \sim P_{\rvc} }\big[\log \big(1 - D(G(\rvz, \rvc), \rvc)\big)\big]
\end{split}
\label{eq:cgan}
\end{equation}
where $\rvc$ denotes a condition and $\rvz$ is noise sampled from a uniform distribution $P_{\rvz}$. In our formulation, $\rvz$ is the latent representation of the input image $\rvx$, learned by the encoder $E(\cdot)$. Finally, the explanation function is defined as,
\begin{equation}
 \mathcal{I}_f(\rvx, \rvc) =  G(E(\rvx), \rvc ).
\end{equation}

For the discriminator in cGAN, we adapted the loss function from the Projection GAN~\citep{Miyato2018CGANsDiscriminator}. The Projection GAN imposes the following structure on the discriminator loss function:
\begin{equation}
    \mathcal{L}_{\text{cGAN}}(D, G)(\rvx, \rvc) := r(\rvx) + r(\rvc|\rvx),
\end{equation}

Here, $r(\rvx)$ is the discriminator logit that  evaluates the visual quality of the generated image. It is the discriminator's attempt to separate real images from the fakes images created by the generator. The second term evaluates the correspondence between the generated image $\rvx_{\rvc}$ and the condition $\rvc$.

\blue{To represent the condition, the discriminator learns an embedding matrix $\mathbf{V}$ with $N$ rows, where $N$ is the number of conditions. The condition is encoded as an  $N$-dimensional one-hot vector which is multiplied by the embedding-matrix to extract the condition-embedding. When $\rvc = n$, the conditional embedding is given as the $n$-th row of the embedding-matrix ($\rvv_n$). The projection is computed as the dot product of the condition-embedding and the features extracted from the fake image,} 
\begin{equation}
\label{eq_cgan}
    \mathcal{L}_{\text{cGAN}}(D, G)(\rvx, \rvc) := r(\rvx) + \rvv_n^T\phi(\rvx),
\end{equation}
 
 \blue{where, $n$ is the current class for the conditional generation and $\phi$ is the feature extractor.}

\blue{ In our use-case, the condition $\rvc$ is the desired posterior probability from the classification function $f$. $\rvc$ is a continuous variable with values in range $[0,1]$. Projection-cGAN requires the condition to be a discrete variable, to be mapped to the embedding matrix $\mathbf{V}$. Hence, we discretize the range $[0,1]$ into $N$ bins, where each bin is one condition. One can view change from $f(\rvx)$ to $\rvc$ as changing the bin index from the current value $C(f(\rvx))$ to $C(\rvc)$ where $C(\cdot)$ returns the bin index. }

\subsection{Classifier consistency}

\blue{Ideally, cGAN should generate a series of smoothly transformed images as we change condition $\rvc$ in range $[0,1]$. These images, when processed by the classifier $f$ should also smoothly change the classification prediction between $[0,1]$. To enforce this, rather than considering bin-index $C(\rvc)$ as a scalar, we consider it as an ordinal-categorical variable, \ie $C(\rvc_1) < C(\rvc_2)$ when $\rvc_1 < \rvc_2$.   Specifically, rather than checking one condition that desired bin-index is equal to some value $n$, $C(\rvc) = n$, we check $n-1$ conditions that desired bin-index is greater than all bin-index which are less than $n$ \ie  $C(\rvc) > i \forall i \in [1, n)$~\citep{Frank2001AClassification}. }

We adapted  Eq.~\ref{eq_cgan} to account for a categorical variable as the condition, by modifying the second term to support ordinal multi-class regression. The modified loss function is as follows:
\begin{equation}
    r(\rvc=n|\rvx) := \sum_{i<n} \rvv_i^{T} \bm{\phi} ( \rvx  ),
\end{equation}

Along with conditional loss for the discriminator, we need additional regularization for the generator to ensure that the actual classifier's outcome, \ie $f(\rvx_{\rvc})$, is very similar to the condition $\rvc$.
To ensure this compatibility with $f$, we further constrain the generator to minimize the KullbackLeibler (KL) divergence that encourages the classifier’s score for $\rvx_{\rvc}$ to be similar to $\rvc$. Our final condition-aware loss is as follows,
\begin{equation}\label{eq:dkl}
\mathcal{L}_f(D,G) := r(\rvc|\rvx) + \KL(f(\rvx_{\rvc}) || \rvc),
\end{equation}
Here, the first term evaluates a conditional probability associated with the generated image given the condition $\rvc$ and is a function of both $G$ and $D$. The second term minimize the KL divergence between the posterior probability for new image $f(\rvx_{\rvc})$ and the desired prediction distribution $\rvc$. It influences only the $G$. Please note, the term $r(\rvx)$ is not appearing in Eq.~\ref{eq:dkl} as it is independent of $\rvc$.

\subsection{Context-aware self consistency}
A valid explanation image is a small modification of the input image, and should preserve the inputs' identity \ie patient-specific information such as the shape of the anatomy or any foreign objects (FO) if present.  While images generated by GAN are shown to be realistic looking~\citep{Karras2019ANetworks}, GAN with an encoder may ignore small or uncommon details in the input image~\citep{Bau2019SeeingGenerate}. To preserve such details, we propose a context-aware reconstruction loss (CARL) that exploits extra information from the input domain to refine the reconstruction results. This additional information comes as semantic segmentation and detection of any FO present in the input image. The CARL is defined as,
\begin{equation}
    \gL_{\text{rec}}(\rvx, \rvx^\prime) = \sum_j \frac{S_j(\rvx) \odot || \rvx -  \rvx^{\prime} ||_1}{\sum S_j(\rvx)} +\KL(O(\rvx) || O(\rvx^{\prime})).
    \label{eq:rec_s}
\end{equation}

Here, $S(\cdot)$ is a pre-trained semantic segmentation network that produces a label map for different regions in the input domain. Rather than minimizing a distance such as $\ell_1$ over the entire image, we minimize the reconstruction loss for each segmentation label ($j$). Such a loss heavily penalizes differences in small regions to enforce local consistency. 

\blue{$O(\rvx)$ is a pre-trained object detector that, given an input image $\rvx$, outputs a number of bounding boxes called region of interests (ROIs). For each bounding box, it outputs 2-d coordinates in the image where the box is located and an associated probability of presence of an object. Using the input image $\rvx$, we obtain the ROIs and associated $O(\rvx)$, which is a probability vector, stating probability of finding an object in each ROI. For reconstructed image $\rvx^{\prime}$, we reuse the ROIs obtained from image $\rvx$ and computed the associated probabilities for the reconstructed image as $O(\rvx^{\prime})$.  Next, we used KL divergence to quantify the difference between probability  vectors as $\KL(O(\rvx) || O(\rvx^{\prime}))$, in eq~\ref{eq:rec_s}.}

\begin{figure}[!ht]
\centering
\includegraphics[width=1.0\linewidth]{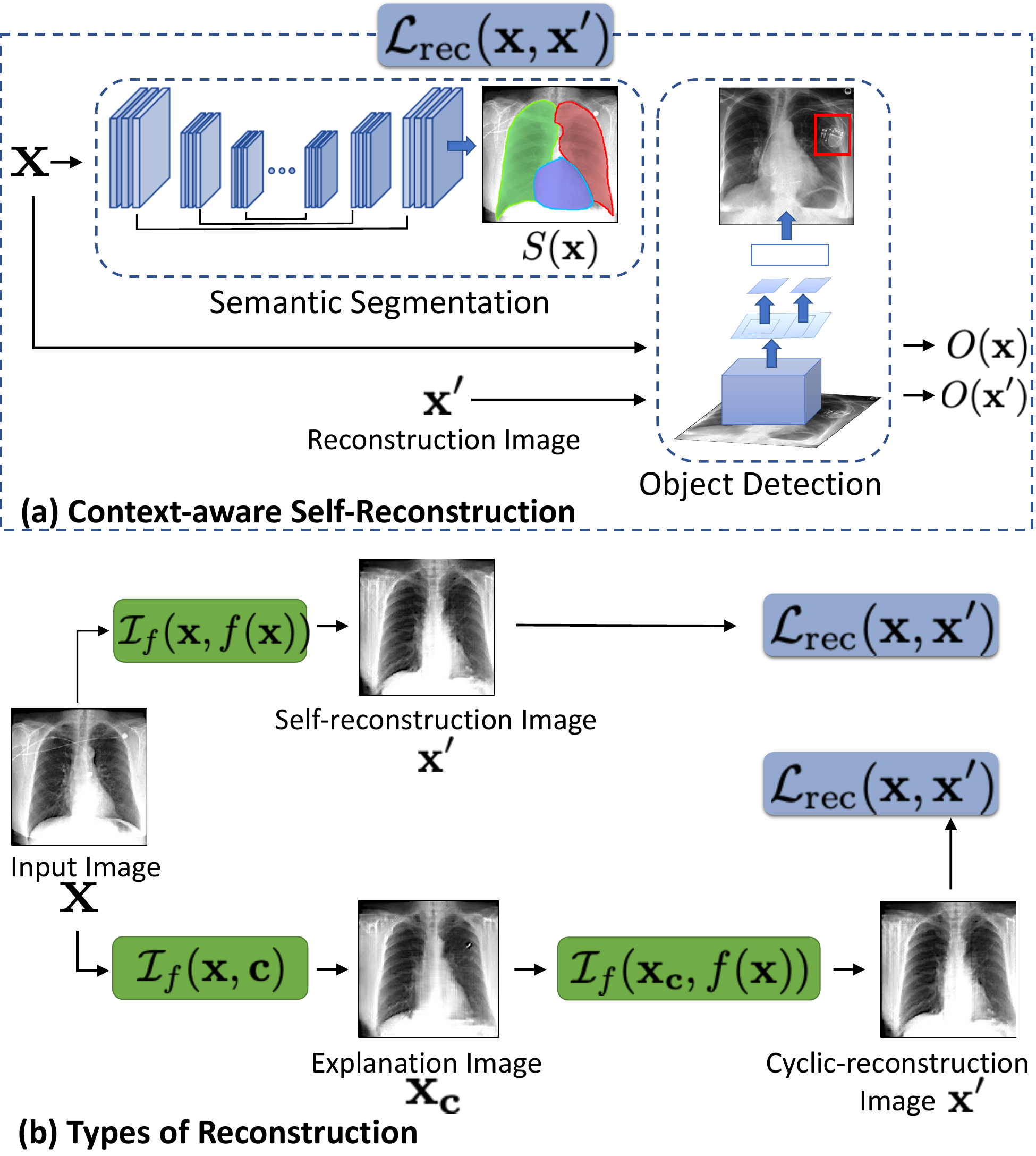}
\caption{(a) A domain-aware self-reconstruction loss with pre-trained semantic segmentation $S(\rvx)$ and object detection $O(\rvx)$ networks. (b) The self and cyclic reconstruction should retain maximum information from $\rvx$. }
\label{Fig_Rec}
\end{figure}

Finally, we used the CAR loss to enforce two essential properties of the explanation function:
\begin{enumerate}
    \item If $\rvc = f(\rvx)$, the self-reconstructed image should resemble the input image.
    \item For $\rvc \neq f(\rvx)$, applying a reverse perturbation on the explanation image $\rvx_{\rvc}$ should recover the initial image \ie $\rvx \approx \mathcal{I}_f(\mathcal{I}_f(\rvx, \rvc), f(\rvx))$.
\end{enumerate}
We enforce these two properties by the following loss,
\begin{equation}
    \gL_{\text{rec}}(E,G) =    \gL_{\text{rec}}(\rvx, \mathcal{I}_f(\rvx, f(\rvx))) +
    \gL_{\text{rec}}(\rvx, \mathcal{I}_f(\mathcal{I}_f(\rvx, \rvc), f(\rvx))).
    \label{eq:rec}
\end{equation}

where $\gL_{\text{rec}}(\cdot)$ is defined in Eq.~\ref{eq:rec_s}.  We minimize this loss only while reconstructing the input image (either by performing self or cyclic reconstruction).  Please note, the classifier $f$ and support networks $S(\cdot)$ and $O(\cdot)$ remained fixed during training.


 \section{Implementation and Evaluation}

\subsection{Dataset}
We performed our experiments on MIMIC-CXR~\cite{Johnson2019MIMIC-CXRReports} dataset consisting of 377K CXR images from 65K patients. The dataset provides  image-level labels for the presence of 14 observations, namely, enlarged cardiomediastinum, cardiomegaly, lung-lesion, lung-opacity, edema, consolidation, pneumonia, atelectasis, pneumothorax, pleural effusion, pleural other, fracture, support devices and no-finding.  

 \subsection{Implementation details}
\blue{ \textbf{Classification model: }We consider classification model that  take as input a single-view chest radiograph and output the probability of each of the 14 observations. Following the prior work on diagnosis classification~\citep{Irvin2019CheXpert:Comparison}, we used DenseNet-121~\citep{Huang2016DenselyNetworks} architecture for
the classifier.  We use the Adam optimizer with default $\beta$-parameters of $\beta_1 = 0.9$, $\beta_2 = 0.999$ and learning
rate $1 \times$ $10^{-4}$ which is fixed for the duration of the training. We used a batch size of 16 images and train for 3 epochs, saving checkpoints every 4800 iterations.} 

The classifier is trained on 198K ($\sim$80\%) frontal view CXR from 51K patients and is test on a held-out set of 50K images from 12K non-overlapping patients. The images are resized to 256 $\times$ 256 and are pre-processed using a standard pipeline involving cropping, re-scaling, and intensity normalization. Our classification model achieved an AUC-ROC of 0.87 for Cardiomegaly, 0.95 for pleural effusion, and 0.91 for edema. These results are comparable to performance of the published model~\citep{Irvin2019CheXpert:Comparison}.

  \textbf{Segmentation network: }Semantic segmentation network $S(\cdot)$ is a 2D U-Net~\citep{Ronneberger2015U-NetSegmentation}  that marks the lung and the heart contour in a CXR. In the absence of ground truth lung and heart segmentation on the MIMIC-CXR dataset, we pre-trained the segmentation network trained on  385 CXRs from Japanese Society of Radiological Technology (JSRT)~\citep{vanGinneken2006SegmentationDatabase} and Montgomery~\citep{Jaeger2014TwoDiseases.} datasets. The pre-trained segmentation network is used in our explanation function to enforce CARL loss and to compute Cardio Thoracic Ratio (CTR). Please refer  SM-Sec.\ref{SM-SS} for details on segmentation network.

 \textbf{Object detector:} We trained a Fast Region-based CNN~\citep{Ren2015FasterNetworks} network as object detector $O(\cdot)$. We trained three independent detectors for three use-cases: detecting foreign objects (FO) such as pacemakers and hardware, detecting healthy costrophenic (CP) recess and detecting blunt CP recess.  
 
 For constructing a training dataset for this object detection, we first collect candidate CXRs for each object by parsing the radiology reports associated with the CXR to find positive mention for \textit{``blunting of the costophrenic angle"} for blunt CP recess, and \emph{``lungs are clear"} for healthy CP recess. For each object, we manually collect bounding box annotations for 300 candidate CXRs.
 

 \textbf{Explanation Function:}
Our explanation function is implemented using TensorFlow version 2.0 and is trained on NVIDID P100 GPU. Before training the explanation function, we assume access to the pre-trained classification function, that we aim to explain. We also assume access to pre-trained segmentation and object detection networks, that are used to enforce CARL loss.

In cGAN, we adapted a ResNet~\citep{He2016DeepRecognition} architecture for the encoder, generator, and discriminator networks. For optimization, we used Adam optimizer~\citep{Kingma2014Adam:Optimization}, with hyper-parameters set to $\alpha = 0.0002, \beta_1 = 0, \beta_2 = 0.9$ and updated the discriminator five times per one update of the generator, and the encoder.

In our experiments, we train three independent explanation functions, for explaining classifier's decision for three class labels; cardiomegaly, pleural effusion (PE), and edema. For training, we divide  $f(\rvx) \in [0,1]$ into $N = 10$ equally-sized bins and trained the cGAN with 10 conditions. To construct the training-set for the explanation function, we randomly sample images from the test-set of the classifier such that each condition (bin-index) have 2500 - 3000 images. Similarly, we created a non-overlapping (unique subjects) evaluation dataset, of 20K images for the explanation function. We created one such dataset for each class label.

\subsection{Evaluation}
\blue{For evaluating the explanations, we randomly sample two groups of real images from the test set of the explanation function (1) a \emph{real-negative} group defined as $\mathcal{X}^{n} = \{\rvx; f(\rvx) < 0.2\}$. It consists of real CXR that are predicted as negative by the classifier $f$ for a target class $k$. (2) A \emph{real-positive} group defined as $\mathcal{X}^{p} = \{\rvx; f(\rvx) > 0.8\}$. For $\mathcal{X}^{n}$, we generated a counterfactual group by setting condition $\rvc = 1.0$ as,  $\mathcal{X}_{\rvc}^{n \rightarrow p} = \{ \mathcal{I}_f(\rvx, \rvc = 1) \forall \rvx \in \mathcal{X}^n\}$. Similarly for $\mathcal{X}^p$, we derived a counterfactual group as   $\mathcal{X}_{\rvc}^{p \rightarrow n} = \{\mathcal{I}_f(\rvx, \rvc = 0) \forall \rvx \in \mathcal{X}^p\}$. We create one such dataset for each target class $k$. Combining the two groups, our set of real images is $\mathcal{X} = \mathcal{X}^n \cup \mathcal{X}^p$ and corresponding set of counterfactual explanations is $\mathcal{X}_{\rvc} = \mathcal{X}_{\rvc}^{n \rightarrow p} \cup \mathcal{X}_{\rvc}^{p \rightarrow n} $. All the results are computed on this evaluation dataset.} 
 
We employ several metrics to quantify different aspects of a valid counterfactual explanation. 

\textbf{Frechet Inception Distance (FID) score:} FID score quantifies the visual similarity between the real images and the synthetic counterfactuals. It computes the distance between the activation distributions as follow,
\begin{equation}
    \text{FID}(\mathcal{X}, \mathcal{X}_{\rvc}) = ||\mu_{\rvx} - \mu_{\rvx_{\rvc}}||_2^{2} + \text{Tr}(\Sigma_{\rvx} + \Sigma_{\rvx_{\rvc}} - 2(\Sigma_{\rvx}\Sigma_{\rvx_{\rvc}})^{\frac{1}{2}}),
\end{equation}
where $\mu$'s and $\Sigma$'s are mean and covariance of the activation vectors derived from the \blue{penultimate layer of a Inception v3 network~\cite{Heusel2017GansEquilibrium} pre-trained on MIMIC-CXR dataset.}

\textbf{Counterfactual Validity (CV) score:}
CV score~\citep{Mothilal2019ExplainingExplanations} is defined as the fraction of counterfactual explanations that successfully flipped the classification decision \ie if the input image is negative, then the explanation is predicted as positive for the target class. CV score is computed as,

\begin{equation}
\text{CV}(\mathcal{X}, \mathcal{X}_{\rvc}) = \frac{\mathbbm{1} (||f(\rvx) - f(\rvx_{\rvc})|| > \tau) }{|\mathcal{X}|}
\end{equation}
where, $\tau$ is the margin between the two prediction distributions. We used $\tau = 0.8$ in our experiments.

\begin{figure*}[!ht]
    \centering
    \includegraphics[width =0.9\textwidth]
    {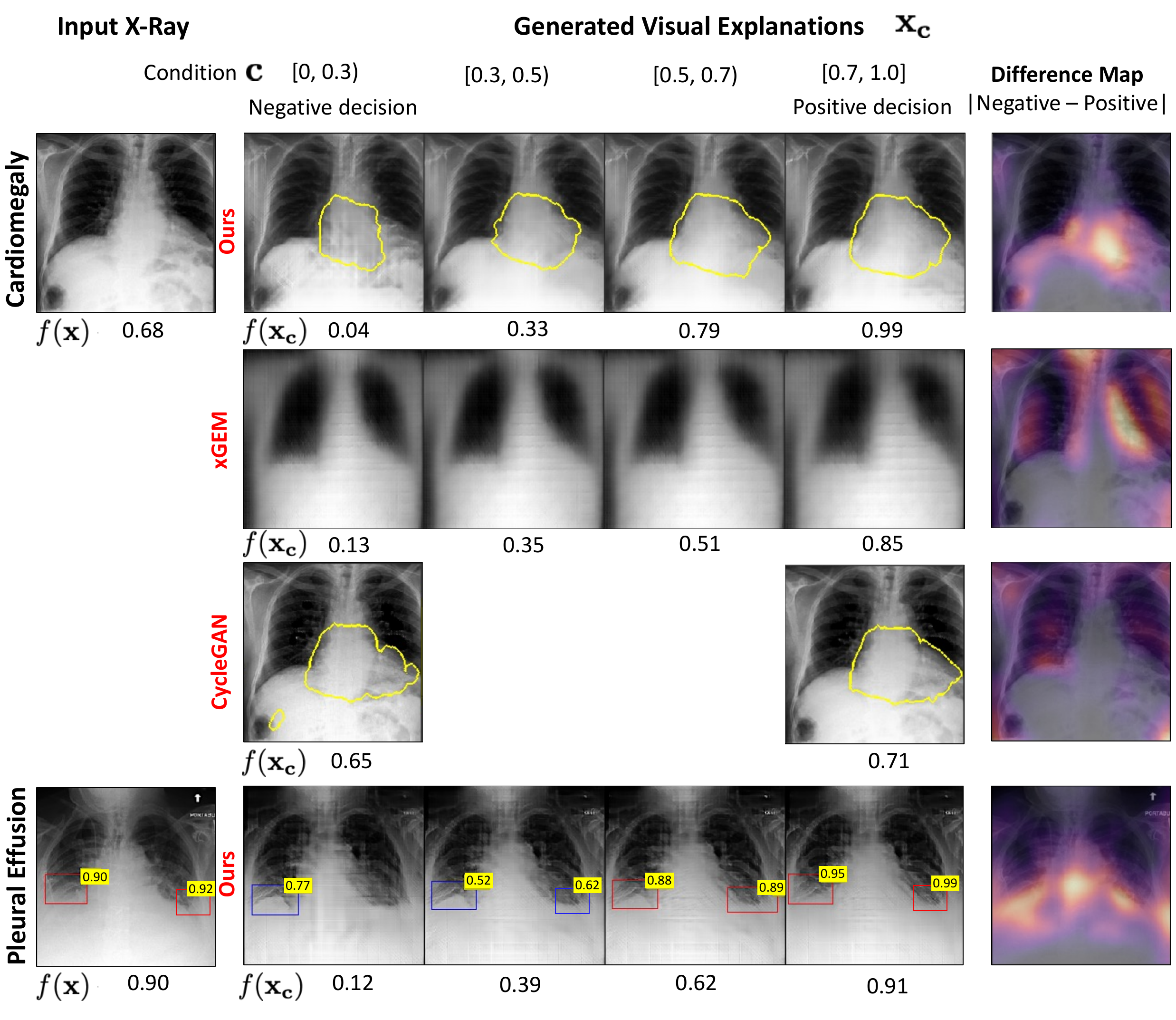}
    \caption{
    Qualitative comparison of the counterfactual explanations generated for two classification tasks, cardiomegaly (first row) and pleural effusion (PE) (last row). The bottom labels are the classifier's predictions for the specific task. For the input image in first column, our model generates a series of images $\rvx_{\rvc}$ as explanations by gradually changing $\rvc$ in range $[0, 1]$. The last column presented a pixel-wise difference map between the explanations at the two extreme ends \ie with condition $\rvc = 0$ (negative decision) and with condition $\rvc = 1$ (positive decision). The heatmap highlights the regions that changed the most during the transformation. For cardiomegaly, we show the heart border in yellow. For PE, we showed the results of an object detector as a bounding-box (BB) over the healthy (blue) and blunt (red) CP recess regions. The number on the top-right of the blue-BB is the Score for detecting a healthy CP recess (SCP). The number on red-BB is 1-SCP.}
    \label{Fig_Quality_Normal}
\end{figure*}

\textbf{Foreign Object Preservation (FOP) score:}
FOP score is the fraction of the real images, with successful detection of FO, in which FO was also detected in the corresponding explanation image $\rvx_{\rvc}$.

\begin{equation}
\text{FOP}(\mathcal{X}, \mathcal{X}_{\rvc}) = \frac{\mathbbm{1} (O(\rvx_{\rvc}) > 0.5) }{|\mathcal{X}|}
\end{equation}
where, $O(\rvx)$ is the probability of finding a FO in image $\rvx$ as predicted by a pre-trained object detector. $O(\rvx) > 0.5 \forall \rvx \in \mathcal{X}$ \ie we  consider images with positive detection of FO in set $\mathcal{X}$.

Next, we define two clinical metrics to quantify the counterfactual changes that leads to the flipping of the classifier's decision. Precisely, we translated the clinical definition of cardiomegaly and pleural effusion into metrics that can be computed using a CXR. 

\textbf{Cardio Thoracic Ratio (CTR): }
We used CTR as the clinical metric to quantify cardiomegaly. CTR is the ratio of the cardiac diameter to the maximum internal diameter of the thoracic cavity. A CTR ratio greater than 0.5 indicates cardiomegaly~\citep{Mensah2015EstablishingScreening,2017EvaluatingEchocardiography,Dimopoulos2013CardiothoracicDisease}. We followed the approach in ~\citep{Chamveha2020AutomatedApproach} to calculate CTR from a CXR.  We use the pre-trained segmentation network $S(\cdot)$ to mark the heart and lung region.
 We calculated heart diameter as the distance between the leftmost and rightmost points from the lung centerline on the heart segmentation. The thoracic diameter is the horizontal distance between the widest points on the lung mask.

\begin{table*}[htpb]
\caption{  The counterfactual validity (CV) score is the fraction of explanations that have an opposite prediction compared to the input image. The FID score quantifies the visual appearance of the explanations. We have normalized the FID scores with respect to the best method (cycleGAN).
}
\label{FID-table}
\begin{center}
\begin{tabular}{c|ccc|ccc|ccc}
\multicolumn{1}{c}{ }  &  \multicolumn{3}{c}{\bf Cardiomegaly} &  \multicolumn{3}{c}{\bf Pleural Effusion}  &  \multicolumn{3}{c}{\bf Edema}\\
& Ours & xGEM & CycleGAN & Ours & xGEM & CycleGAN & Ours & xGEM & CycleGAN\\
\hline
\multicolumn{10}{c}{\bf Counterfactual Validity Score} \\
\hline
$\text{CV}(\mathcal{X}, \mathcal{X}_{\rvc})$& \bf 0.91&  \bf 0.91& 0.43 & \bf 0.97 & \bf 0.97& 0.49 & \bf 0.98 & 0.66& 0.57\\
\hline
\hline
\multicolumn{10}{c}{\bf FID score} \\
\hline
Negative group: FID($\mathcal{X}^n, \mathcal{X}_{\rvc}^{n \rightarrow p}$)& 4.3 & 12.5& \bf 1 & 3.7 & 9.2  & \bf 1 & 1.9 & 5.0& \bf 1\\
Positive group: FID($\mathcal{X}^p, \mathcal{X}_{\rvc}^{p \rightarrow n}$) & 2.4 & 5.6 &\bf 1 & 3.4 & 10.1& \bf1 & 1.3 & 3.5 & \bf 1\\
\hline
\end{tabular}
\end{center}
\end{table*}

 \textbf{Score for detecting a healthy Costophrenic recess (SCP): }We first identify CP recess in a CXR and then classify it as healthy or blunt to quantify pleural effusion. The fluid accumulation in CP recess may lead to the diaphragm's flattening and the associated blunting of the angle between the chest wall and the diaphragm arc, called costophrenic angle (CPA). The blunt CPA is an indication of pleural effusion~\citep{Maduskar2016AutomaticRadiographs,PleuralTomography}. Marking the CPA angle on a CXR requires expert supervision, while annotating the CP region with a bounding box is a much simpler task (\emph{see} SM-Fig.~\ref{FIG_CPA}).  We learned an object detector to identify healthy or blunt CP recess in the CXRs and used SCP as our evaluation metric.

\section{Experiments and Results}

We performed four sets of experiments on CXR dataset:

(1) In Section~\ref{Desiderata}, we evaluated the validity of our counterfactual explanations and compared them against xGEM~\citep{Joshi2018XGEMs:Models} and CycleGAN~\citep{Narayanaswamy2020ScientificTranslation,DeGrave2020AISignal}. 

(2) In Section~\ref{sm},  we compared against the saliency-based methods to provide \emph{post-hoc} model explanation. 

(3) In Section~\ref{dse}, we associate the counterfactual changes in our explanation with  the clinical definitions of two diagnosis, cardiomegaly and pleural effusion.

(4) In Section~\ref{he}, we present a  clinical study that collects subjective feedback from radiology residents on three different explanation approaches, saliency maps, cycleGAN and ours.

\subsection{Validity of counterfactual explanations}
\label{Desiderata}

A valid counterfactual explanation resembles the query image while having perceivable differences that achieves an opposite classification decision as compared to the query image from the classifier. In Fig.~\ref{Fig_Quality_Normal}, we present qualitative examples of counterfactual explanations from our method and compared them against those obtained from xGEM and CycleGAN.

\subsubsection{Classifier consistency}
In Fig.~\ref{Fig_Quality_Normal}, we observe that the explanation images gradually flip their decision $f(\rvx_{\rvc})$ (bottom label) as we go from left to right. Table.~\ref{FID-table} summarizes our results on CV score metric. A high CV score for our model confirms that the condition used in cGAN is successfully met and the generated explanations are successfully flipping the classification decision and hence, are consistent with the classifier. 

On the other hand, cycleGAN achieves a CV score of about 50\%, thus creating valid counterfactual images only half of the times. In a deployment scenario, a counterfactual explanation that fails to flip the classification decision would be rejected as an invalid, and hence half of the explanations provided by cycleGAN would be rejected.

 

\blue{Our formulation constraints the condition $\rvc$ to vary linearly with the actual prediction  $f(\rvx_{\rvc})$ \ie if we increase $\rvc$ in range $[0,1]$ then the cGAN should create an image $\rvx_{\rvc}$ such that condition $\rvc$ is met and the actual prediction $f(\rvx_{\rvc})$ should also increase. Further, consider a scenario when $\rvc = 1.0$. The expected behaviour is $f(\rvx_{\rvc=1.0}) \approx 1.0$ and also, $f(\rvx_{\rvc = 1.0}) > f(\rvx_{\rvc = 0.9}) >  f(\rvx_{\rvc = 0.8})> \cdots > 0$, where different $\rvx_{\rvc}$ are generated using same $\rvx$ but different conditions. }
 
\begin{figure}[!ht]
    \centering
    \includegraphics[width = 0.95\linewidth]
    {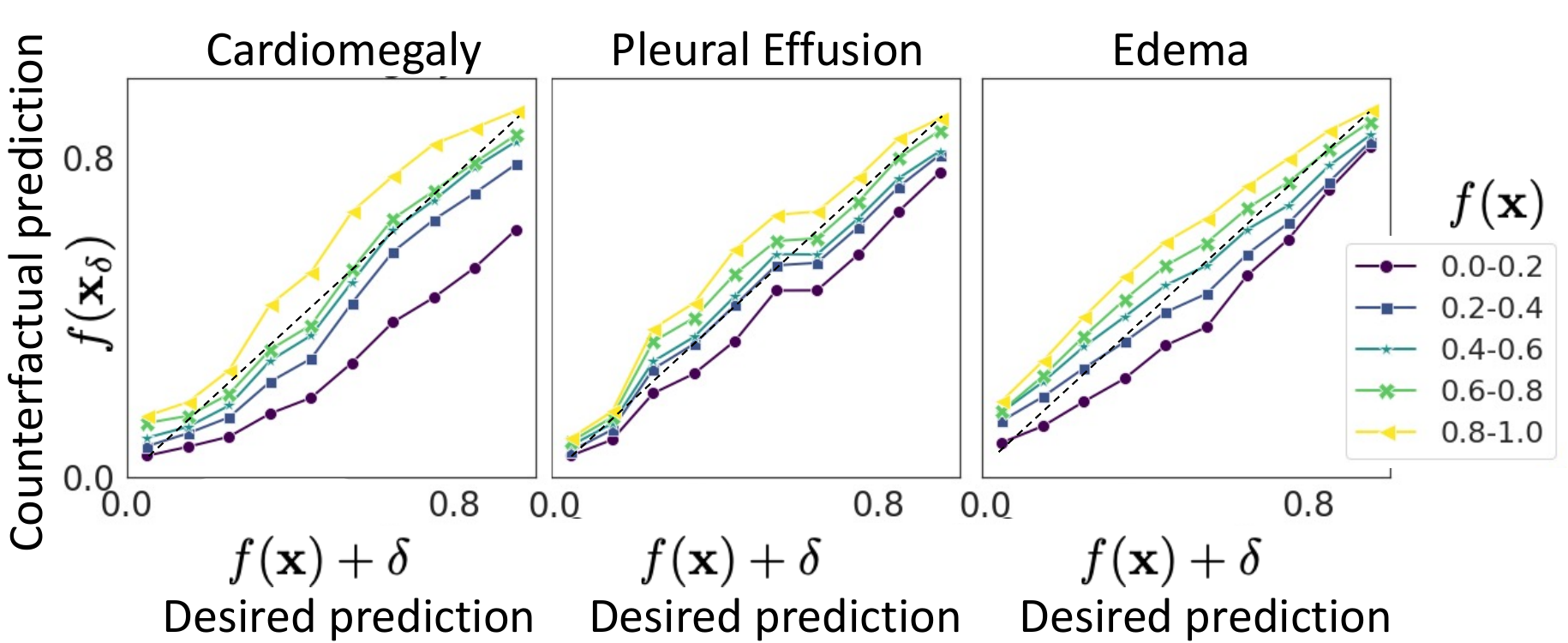}
    \caption{ The plot of condition, $\rvc$ (desired prediction), against actual response of the classifier on generated explanations, $f(\rvx_{\rvc})$. Each line represents a set of input images with prediction $f(\rvx)$ in a given range. Plots for xGEM and cycleGAN are shown in SM-Fig.~\ref{Fig_Cls_ext}. }
    \label{Fig_Cls}
\end{figure}

 To verify this behaviour, we group images in the test-set of the explanation function, into five non-overlapping groups based on their original prediction $f(\rvx)$. Next, for each image, we created 10 explanation images by discretising the range $[0,1]$ into 10 bins. In Fig~\ref{Fig_Cls}, we represented each input group as a line and plotted the average response of the classifier \ie $f(\rvx_{\rvc})$ for explanations generated with a same condition $\rvc$.   The positive slope of the line-plot, parallel to $y = x$ line confirms that starting from images with low $f(\rvx)$, our model creates fake images such that $f(\rvx_{\rvc})$ is high and vice-versa.

\subsubsection{Visual quality} 
Qualitatively, the counterfactual explanations generated by our method look visually similar to the query image (\emph{see} Fig.~\ref{Fig_Quality_Normal}).  Table.~\ref{FID-table} reports the FID score for our method and compares them against  xGEM and cycleGAN. Our approach achieved a lower FID score as compared to xGEM. xGEM has very high FID score, thus creating counterfactual images that are very different from the query image and hence are unsuitable for deployment. CycleGAN achieved the least FID score, thus generating the most realistic images as explanations.

\subsubsection{Identity preservation} 
Ideally, a counterfactual explanation should differ in semantic features associated with the target task while retaining unique properties of the input, such as foreign objects (FOs). FO provide critical information to identify the patient in an x-ray. The disappearance of FO in explanation images may create confusion that explanation images show a different patient.

\begin{table}[!ht]
\centering
\caption{The foreign object preservation (FOP) score for our model with and without the context-aware reconstruction loss (CARL). FOP score depends on the performance of FO detector.}
\label{OD-table}
\begin{tabular}{c|c|cc}
Foreign &  \multicolumn{2}{c}{\bf FOP score}\\
 Object (FO) &  Ours with CARL & Ours with $\ell_1$ \\
\hline
Pacemaker  & \bf 0.52 & 0.40   \\
Hardware  & \bf 0.63& 0.32   \\
\hline
\end{tabular}
\end{table}

\begin{figure}[h]
    \centering
    \includegraphics[width = 0.8\linewidth]
    {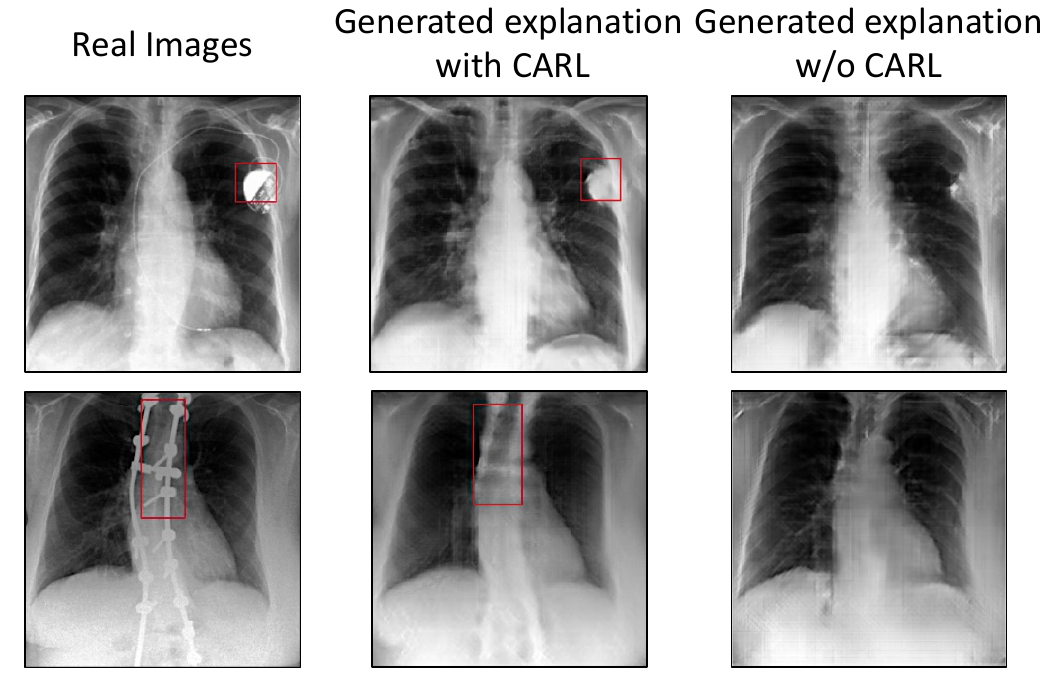}
    \caption{Fidelity of generated images with respect to preserving FO.}
   \label{Fig_Pacemaker}
\end{figure}

In this experiment, we quantify the strength of our revised CARL loss in preserving FO in explanation images compared to an image-level $\ell_1$ reconstruction loss. In Table~\ref{OD-table}, we report the results on the FOP score metric. Our model with CARL obtained a higher FOP score. The FO detector network has an accuracy of 80\%.
Fig.~\ref{Fig_Pacemaker} presents examples of counterfactual explanations generated by our model with and without the CARL. Our results confirm that CARL is an improvement over $\ell_1$ reconstruction loss. We further provide a detailed ablation study over different components of our loss in SM-Sec.\ref{SM-AS}.

\subsection{Comparison with saliency maps}
\label{sm}

Popular existing approaches for post-hoc model explanation includes explaining using a  saliency-map~\citep{Pasa2019EfficientVisualization,Irvin2019CheXpert:Comparison}. \blue{To compare against such methods,  we approximated  a saliency map as a pixel-wise difference map between the explanations at the two extreme ends \ie with condition $\rvc = 0$ (negative decision) and with condition $\rvc = 1$ (positive decision).} For proper comparison, we normalized the absolute values of the saliency maps in the range $[0, 1]$.

\begin{figure*}[ht]
\centering
\includegraphics[width=0.8\linewidth]{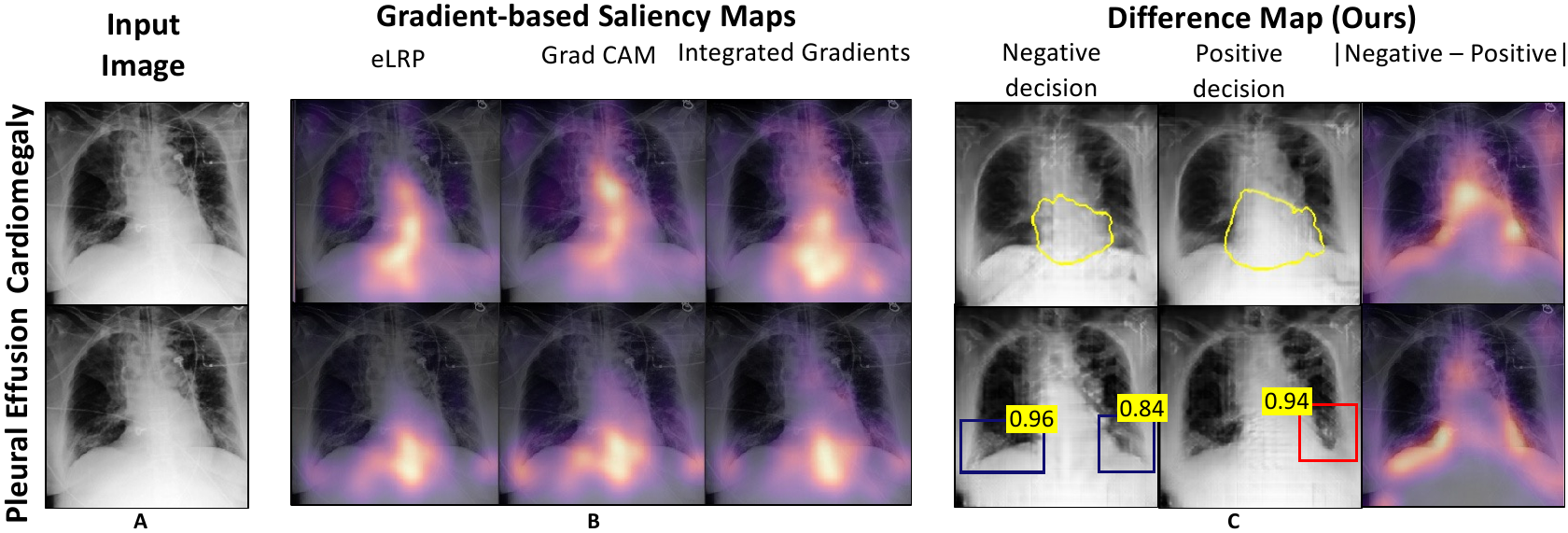}
\caption{Comparison of our method against different gradient-based methods. A: Input image; B: Saliency maps from existing works; C: Our simulation of saliency map as difference map between the normal and abnormal explanation images. More examples are shown in SM-Fig.~\ref{Fig_c}.}
\label{Fig_saliency}
\end{figure*}

 In clinical setting, multiple diagnosis may affect the same anatomical region. In this case, the saliency map may highlight same regions as important for multiple target tasks. Fig.~\ref{Fig_saliency} is showing one such example. Our method not only provides a saliency map, but also counterfactual images to demonstrate how image features in those relevant regions should be modify to change the classification decision.  
 
\begin{figure}[h]
    \centering
    \includegraphics[width = 0.95\linewidth]
    {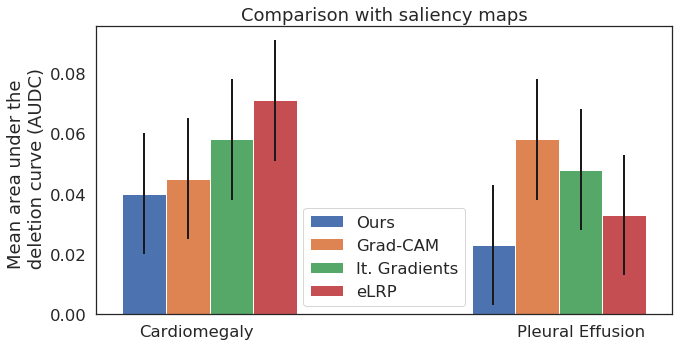}
    \caption{Quantitative comparison of our method against gradient-based methods. Mean area under the deletion curve (AUDC), plotted as a function of the fraction of removed pixels. A low AUDC shows a sharp drop in prediction accuracy as more pixels are deleted.}
   \label{fig_sm}
\end{figure}
 
\emph{Quantitatively evaluation:} In this experiment, we quantitatively compare different methods for generating saliency maps, to show that important regions identified by these methods are actually relevant for classification decision. Specifically, we used the \textit{deletion} evaluation metric~\citep{Petsiuk2018RISE:Models,removeFirst}. \blue{For each image in set $\mathcal{X}^p$, we derived saliency maps using different methods. We used the saliency information to sort the pixels based on their importance. Next, we gradually removed top $x\%$ of important pixels by selectively impainting the removed region based on its surroundings. We processed the resulting image through the classifier and measure the output probability. We repeated this process while gradually increasing the fraction of removed pixels.}

\begin{figure*}[h!]
\centering
\includegraphics[width = 0.9\textwidth]
{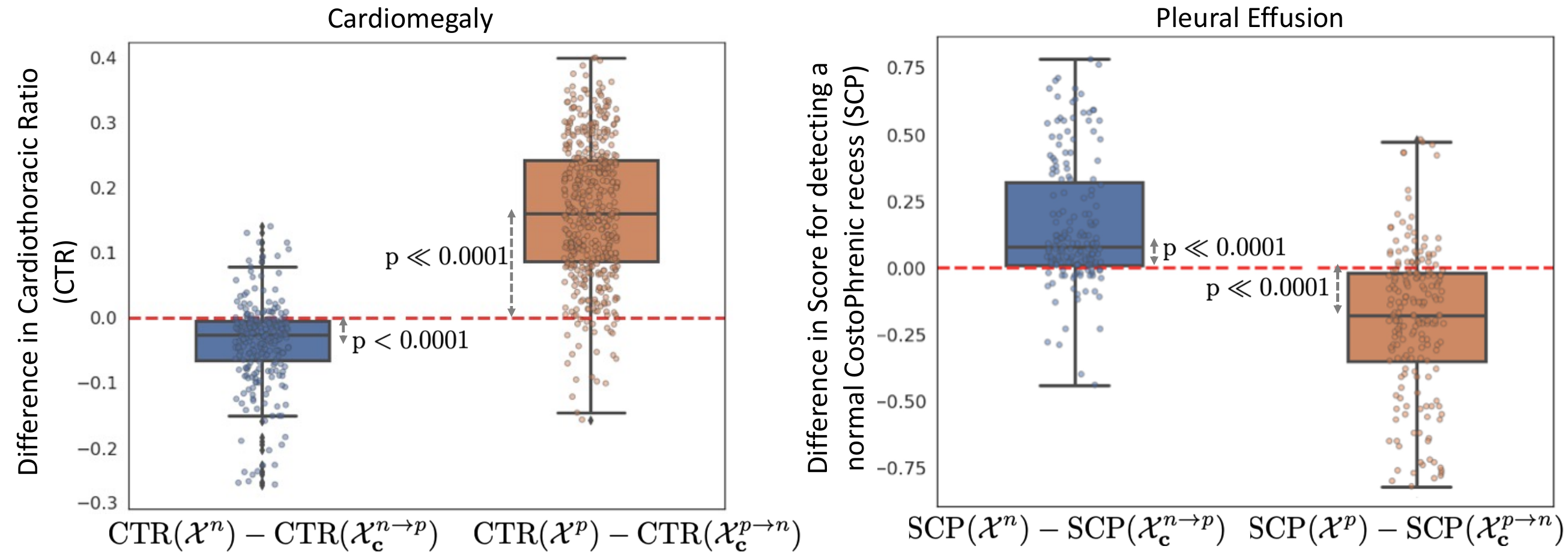}
\caption{Box plots to show distributions of pairwise differences in clinical metrics, CTR for cardiomegaly and the Score of normal CP recess (SCP) for pleural effusion, before (real) and after (counterfactual) our generative counterfactual creation process. The mean value corresponds to the average causal effect of the clinical-metric on the target task. The low p-values for the dependent t-test statistics confirm the statistically significant difference in the distributions of metrics for real and counterfactual images. The mean and standard deviation for the statistic tests are summarized in SM-Table~\ref{t-table}.} 
\label{Fig_Box}
\end{figure*}

\blue{For each image, we plotted the updated classification probability as a function of the fraction of removed pixels, to obtain the deletion curve and measure its area under the deletion curve (AUDC). A sharp decline in classification probability shows that the removed pixels were actually important for classification decision. A sharp decline results in smaller AUDC, and demonstrates the high sensitivity of the classifier in the salient regions. In Fig.~\ref{fig_sm}, we reported the mean and standard deviation of the AUDC metric over the set $\mathcal{X}^p$.
Our method achieved the lowest AUDC, confirming the high sensitivity of the classifier in the salient regions identified by our method.}

\subsection{Disease-specific evaluation}
\label{dse}
In this experiment, we demonstrated the clinical relevance of our explanations. We defined two clinical metrics, cardiothoracic ratio (CTR) for cardiomegaly and score of the normal costophrenic recess (SCP) for pleural effusion.
We used these metrics to quantify the counterfactual changes between normal (negative diagnosis) and abnormal (positive diagnosis) populations, as identified by the given classifier. If the change in classification decision is associated with the corresponding change in clinical-metric, we can conclude that the classifier considers clinically relevant information in its diagnosis prediction.

 We conducted a paired t-test to determine the effect of counterfactual perturbation on the clinical metric for the respective diagnosis. To perform the test, we considered the two groups of real images $\mathcal{X}^n$ and $\mathcal{X}^p$ and their corresponding counterfactual groups $\mathcal{X}_{\rvc}^{n \rightarrow p}$ and $\mathcal{X}_{\rvc}^{p \rightarrow n}$ respectively. In Fig.~\ref{Fig_Box}, we showed the distribution of differences in CTR for cardiomegaly and SCP for PE in a pair-wise comparison between real images and their respective counterfactuals. Patients with cardiomegaly have higher CTR as compared to normal subjects.  Consistent with clinical knowledge, in Fig.~\ref{Fig_Box}, we observe a negative mean difference for CTR($\mathcal{X}^n$) $-$ CTR($\mathcal{X}_{cf}^{n\rightarrow p}$) (a p-value of $< 0.0001$) and a positive mean difference for CTR($\mathcal{X}^a$) $-$ CTR($\mathcal{X}_{cf}^n$) (with a p-value of $\ll 0.0001$). The low p-value in the dependent t-test statistics supports the alternate hypothesis that the difference in the two groups is statistically significant, and this difference is unlikely to be caused by sampling error or by chance.

By design, the object detector assigns a high SCP to a healthy CP recess with no evidence of blunting CPW. Consistent with our expectation, we observe a positive mean difference for SCP($\mathcal{X}^n$) $-$ SCP($\mathcal{X}_{\rvc}^{n \rightarrow p}$) (with a p-value of $\ll 0.0001$) and a negative mean difference for SCP($\mathcal{X}^p$) $-$ SCP($\mathcal{X}_{\rvc}^{p \rightarrow n}$) (with a p-value of $\ll 0.0001$). A low p-value confirmed the statistically significant difference in SCP for real images and their corresponding counterfactuals.

\subsection{Human evaluation}
\label{he}
\blue{We conducted a human-grounded experiment with diagnostic radiology residents to compare different styles of explanations (no explanation, saliency map, cycleGAN explanation, and our counterfactual explanation) by evaluating different aspects of explanations: (1) understandability,  (2) classifier's decision justification, (3) visual quality, (d) identity preservation, and (5) overall helpfulness of an explanation to the users. }

\blue{Our results show that our counterfactual explanation  was the only explanation method that significantly improved  the users' understanding of the classifier's decision compared to the no-explanation baseline. In addition, our counterfactual explanation had a significantly higher classifier's decision justification than the cycleGAN explanation, indicating that the participants found a good evidence for the classifier's decision more frequently in our counterfactual explanation as compared to cycleGAN explanation. }

\blue{Further, cycleGan explanation performed better in terms of visual quality and identity preservation. However, at times the cycleGAN explanations were identical to the query image, thus providing inconclusive explanations.
Overall the participants found our explanation method the most helpful method in understanding the assessment made by the AI system in comparison to other explanation methods. 
Below, we describe the design of the study, the data analysis methods, along with the results of the experiment in detail.}
\subsubsection{Experiment Design}
\blue{We conducted an online survey experiment with 12 diagnostic radiology residents. Participants first reviewed an instruction script, which described the AI system developed to provide an autonomous diagnosis for CXR findings such as cardiomegaly. The study comprised of the radiologists evaluating six CXR images which were presented in random order to them. For selecting these siz CXR, we first, divided the  test-set of the explanation function for cardiomegaly  into three groups, positive ($f(\rvx) \in [0.8, 1.0]$), mild ($f(\rvx) \in [0.4,0.6]$) and negative ($f(\rvx) \in [0.0, 0.2]$). Next, we randomly selected two CXR images from each group. The six CXR images were anonymized as part of the MIMIC-CXR dataset protocol. }

\blue{For each image, we had the same procedure consisted of a diagnosis tasks, followed by four explanation conditions, and ended by a final evaluation question between the explanation conditions. Further details of the study design are includes in SM-Section~\ref{sm-he}.} 

\blue{\emph{Diagnosis:}
For each CXR image, we first asked a participant to provide their diagnosis for cardiomegaly. This question ensures that the participants carefully consider the imaging features that helped them diagnose. Subsequently, the participants were presented with the classifier's decision and were asked to provide feedback on whether they agreed. }

\blue{\emph{Explanation Conditions:}
Next, the study provides the classifier's decision  with the following explanation conditions: 
\begin{enumerate}
    \item \textbf{No explanation (Baseline)}: This condition simply provides the classifier decision without any explanation, and is used as the control condition.
    \item \textbf{Saliency map}: A heat map overlaid on the query CXR, highlighting essential regions for the classifier's decision.
    \item \textbf{CycleGAN explanation}: A video loop over two CXR images, corresponding to the query CXR transformation with a negative and a positive decision for cardiomegaly.
    \item \textbf{Our counterfactual explanation}: A video showing a series of CXR images gradually changing the classifier's decision from negative to positive.
\end{enumerate}}

\blue{Please note that after showing the baseline condition, we provided the other explanation conditions in random order to avoid any learning or biasing effects. }

\blue{\emph{Evaluation metrics:} Given the classifier's decision and corresponding explanation, we consider the following metrics to compare different explanation conditions:}

\begin{enumerate}

\item \blue{\textbf{Understandability}: For each explanation condition, the study included a question to measure whether the end-user understood the classifier's decision, when explanation was provided. The participants were asked to rate agreement with \emph{``I understand how the AI system made the above assessment for Cardiomegaly"}.}
\item \blue{\textbf{Classifier's decision justification}: Human user's may perceive explanations as the reason for the classifier's decision. For the cycleGAN and our counterfactual explanation conditions, we quantify whether the provided explanation were actually related to the classification task by measuring  the participants' agreement with \emph{``The changes in the video are related to Cardiomegaly"}.}
\item \blue{\textbf{Visual quality}: The study quantifies the proximity between the explanation images and the query CXR by measuring the participants' agreement with "\emph{Images in the video look like a chest x-ray.}".}

\item \blue{\textbf{Identity preservation}: The study also measures the extent to which participants think the explanation images correspond to the same subject as the query CXR by measuring the participants' agreement with \emph{``Images in the video look like the chest x-ray from a given subject"}.}

\item \blue{{\textbf{Helpfulness:}} For each CXR image, we asked the participants to select the most helpful explanation condition in understanding the classifier's decision, \emph{``Which explanation helped you the most in understanding the assessment made by the AI system?"}. This evaluation metric directly compares the different explanation conditions. }
\end{enumerate}
 \blue{All metrics, but the helpfulness metric were evaluated for agreement  on a 5-point Likert scale, where one means ``\emph{strongly disagree}" and five means ``\emph{strongly agree}". }

\blue{\emph{Free-form Response:} After each question, we also asked the participants a free-form question: ``\emph{Please explain your selection in a few words.}" We used answers to these questions to triangulate our findings and complement our quantitative metrics by understanding our participants' thought-processes and reasoning.} 

\blue{\emph{Participants.} Our participants include 12 diagnostic radiology residents who have completed medical school and have been in the residency program for one or more years. On average, the participants finished the survey in 40 minutes and were paid \$100 for their participation in the study.}

\subsubsection{Data analysis} 
\blue{For each evaluation metric, the study asked the same question to the participants while showing them different explanations. For each question, we gather 72 responses (6 - number of CXR images $\times$ 12 - number of participants). }

\blue{For the understandability and helpfulness metrics, we conducted a one-way ANOVA test to determine if there is a statistically significant difference between the mean metric scores for the four explanation conditions. Specifically, we built a one-way ANOVA with the metric as our dependent variable and analyzed agreement rating as the independent variable. If we found a significant difference in the ANOVA method, we ran Tukey’s Honestly Significant Difference (HSD) posthoc test to perform a pair-wise comparison between different explanation conditions. }

\blue{We measured the classifier's  decision justification,  visual quality and identity preservation metrics only for the cycleGAN and our counterfactual explanations. We conducted  paired t-tests to compare these evaluation metrics between these two explanation conditions.  We also qualitatively analyzed the participants' free-form responses to find themes and patterns in their responses. }

\subsubsection{Results}
\begin{figure}[h]
    \centering
    \includegraphics[width = 0.95\linewidth]
    {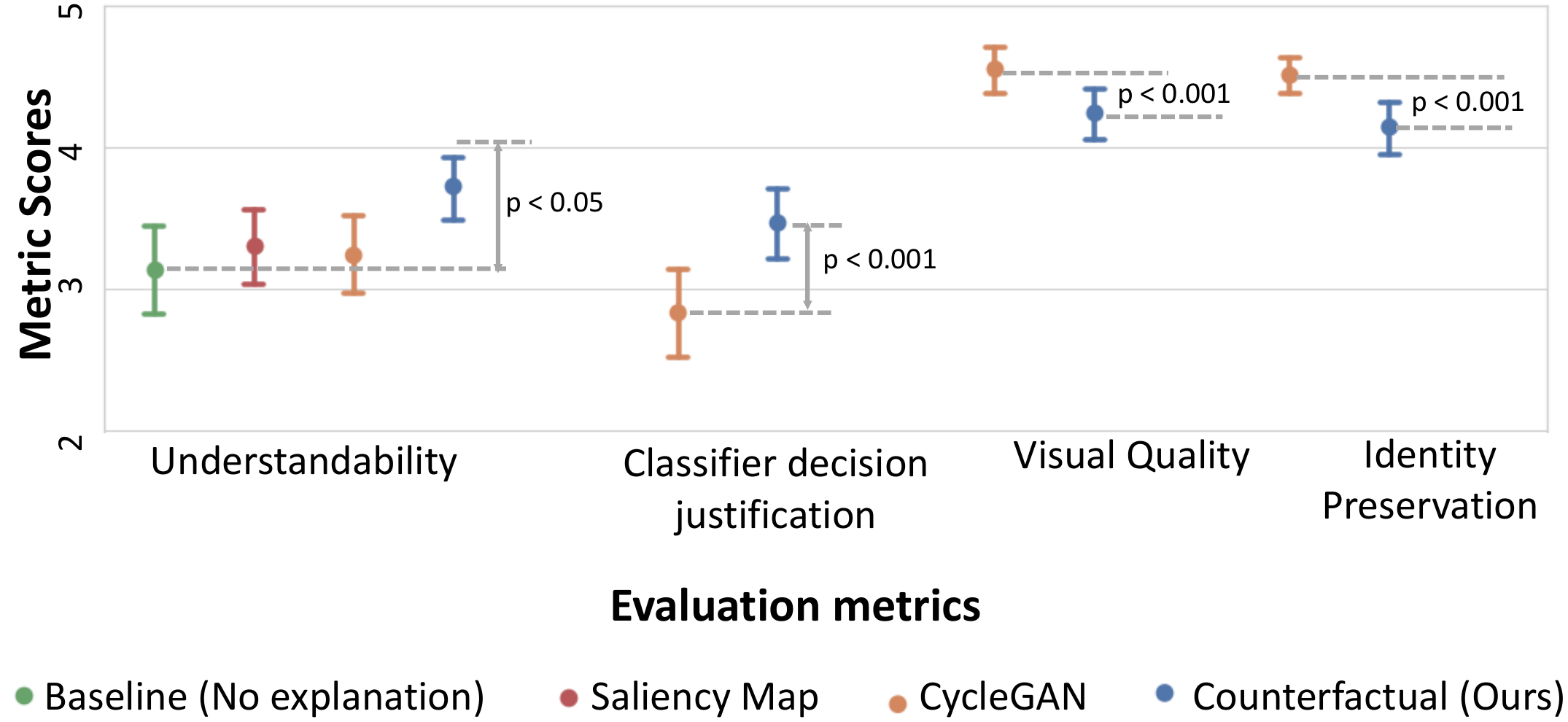}
    \caption{Comparing the evaluation metrics of understandability, classifier's decision justification, visual quality, and identity preservation across the different explanation conditions.}
   \label{fig_he}
\end{figure}

\blue{Fig.~\ref{fig_he} shows the mean score for the evaluation metrics of understandability, classifier's decision justification, visual quality, and identity preservation among the different explanation conditions. Below, we report the statistical analysis for these results, followed by analysis of the participants' free-form responses to understand the reasons behind these results.}

\blue{\emph{Understandability:} The results show that our counterfactual explanation was the most understandable explanation to the participants. A one-way ANOVA revealed that there was a statistically significant difference in the understandability metric between at least two explanation conditions (F(3, 284) = [3.39], p=0.019). The Tukey post-hoc test showed that the  understandability metric for our counterfactual explanation was significantly higher than the no-explanation baseline (p = 0.018). However, there was no statistically significant difference in mean scores between other pairs of explanations (refer to Table~\ref{table_anova}, ``Understandability" column). 
This finding indicates that providing our counterfactual explanations along with the classifier's decision made the algorithm most understandable to our clinical participants,  while other explanation conditions, 
 saliency map and cycleGAN failed to achieve significant difference from no-explanation baseline on the understandability metric. Next, we use responses from free-text question to supplement our findings.}

\blue{For the no-explanation baseline, the primary reason for poor understanding was the absence of explanation (n=30), (\eg  they stated that \emph{``there is no indication as to how the AI made this decision"}). Interestingly, many responses (n=23) either associated their high understanding with the correct classification decision \ie participants understood the decision as the decision is correct (\emph{``I agree, it is small and normal"}) or they assumed the AI-system is using similar reasoning as them to arrive at its decision (\emph{``I assume the AI is just measuring the width of the heart compared to the thorax", ``Assume the AI measured the CT ratio and diagnosed accordingly."}).}

\blue{Participants' mostly found saliency maps to be correct but incomplete  (n=23), (\emph{``Unclear how assessment can be made without including additional regions"}). Specifically, for cardiomegaly, the saliency maps were highlighting parts of the heart and not its border (\emph{``Not sure how it gauges not looking at the border"}) or thoracic diameter (\emph{``thoracic diameter cannot be assessed using highlighted regions of heat map"}). We observe a similar result in Fig.~\ref{Fig_saliency}, where the heatmap focuses on the heart but not its border. Further, some participants expressed a concern that they didn't understand how relevant regions were used to derive the decision (\emph{``i understand where it examined but not how that means definite cardiomegaly"}).}

\blue{For cycleGAN explanation, the primary reason for poor understanding was the minimal perceptible changes between the negative and positive images (n=3), (\emph{``There is no change in the video."}). In contrast, many participant's explicitly reported an improved understanding of the classifier's decision in the presence of our counterfactual explanations (n=33), (\emph{``I think the AI looking at the borders makes sense.", ``i can better understand what the AI is picking up on with the progression video"}).}

 \blue{\emph{Classifier's decision justification}:
 Our counterfactual explanation (M=3.46; SD=1.12) achieved a positive mean difference of 0.63 on this metric as compared to cycleGAN (M=2.83; SD=1.33), with t(71)=3.55 and $p<0.001$. This result indicates that the  participants found a good evidence for the predicted class (cardiomegaly), much frequently  in our counterfactual explanations as compared to cycleGAN.} 
 

  \blue{Most responses (n=25) explicitly mentioned visualizing changes related to cardiomegaly such as an enlarged heart in our explanation video as compared to cycleGAN (n=17). In cycleGAN, many reported that changes in the explanation video was not perceptible (n=23). Further, the participants reported changes in density, windowing level or other attributes which were not related to cardiomegaly (\emph{``Decreasing the density does not impact how I assess for cardiomegaly.", ``they could be or just secondary to windowing the radiograph"}). Such responses were observed in both cycleGAN (n=17) and our explanation (n=17). This indicates that the classifier may have associated such secondary information (short-cuts) with cardiomegaly diagnosis. A more in-depth analysis is required to quantify the classifiers' behaviour. }


\blue{\emph{Visual quality and identity preservation}: We observe a negative mean difference of 0.31 and 0.37 between our and cycleGAN explanation methods in visual quality and identity preservation metrics, respectively. The mean score for visual quality was higher for cycleGAN (M=4.55; SD=0.71) as compared to our method (M=4.24; SD=0.80) with t(71)=3.49 and $p<0.001$. Similarly, the mean  score for identity preservation was also higher for cycleGAN (M=4.51; SD=0.56) as compared to our method (M=4.14; SD=0.78) with t(71)=3.96 and $p<0.001$.}

\blue{Most of the responses (n=69) agreed that the CycleGAN explanation were marked as highly similar to the query CXR image. These results are consistent with our earlier results, that cycleGAN has better visual quality with a lower FID score (\emph{see} Table.~\ref{FID-table}). However, in some responses, the participants pointed out that the explanation images were almost identical to the query image (\emph{``There's virtually no differences. This is within the spectrum of a repeat chest x-ray for instance."}). An explanation image identical to the query image provides no information about the classifier's decision.  Further, similar looking CXR will also result in similar classification decision, and hence will fail to flip the classification decision. As a result, we also observed a lower agreement in the classifier consistency metric and a lower counterfactual validity score in Table.~\ref{FID-table} for cycleGAN.}

\blue{\emph{Helpfulness:} In our concluding question, \emph{``Which explanation helped you the most in understanding the assessment made by the AI system?"}, ~57\% of the responses selected our counterfactual explanation as the most helpful method. A one-way ANOVA revealed that there was a statistically significant difference in the helpfulness metric between at least two explanation conditions (F(3, 284) = [21.5], $p<0.0001$). In pair-wise Tukey's HSD posthoc test,  we found that the mean  usefulness metric for our counterfactual explanations was significantly different from all the rest explanation conditions($p<0.0001$). Table~\ref{table_anova} ( ``Helpfulness" column) summarizes these results.}

\blue{These results indicates that the participant's selected our counterfactual explanations as the most helpful form of explanation for understanding the classifier's decision.}

\begin{table}[h]
\centering
\caption{ Results for one-way ANOVA for understandability metric, followed by Tukey's HSD post-hoc test between different levels of agreement. Note that the mean value for E4 (our counterfactual explanation) is the highest, indicating that our explanations helped users the most in understanding the classifier's decision. *$p < 0.05$; ***$p<0.0001$  }
\label{table_anova}
\small
\begin{tabular}{c|c|c|c|c|c}
\multicolumn{3}{c|}{\bf Understandability} & \multicolumn{3}{c}{\bf Helpfulness} \\
\multicolumn{3}{c|}{F(3, 284) = 3.39} & \multicolumn{3}{c}{F(3, 284) = 21.5}\\
\multicolumn{3}{c|}{$p < 0.05$} & \multicolumn{3}{c}{$p < 0.001$}\\
\hline
\multicolumn{2}{c|}{ Explanation method}  &  $p$ & \multicolumn{2}{c|}{ Explanation method}  &  $p$\\
\hline
E1 (No explanation) & E2 & & E1 & E2 & \\
M=3.14  & E3 &  & M=0.05 & E3 & \\
SD=1.39 & E4 & * & SD=0.23 & E4 & ***\\
\hline
E2 (Saliency Map) & E1 & & E2 & E1 & \\
M=3.31 & E3 & & M=0.18 & E3 & \\
SD=1.13 & E4 & & SD=0.39& E4& *** \\
\hline
E3 (CycleGAN) & E1 & & E3 & E1&\\
M=3.24 & E2 & &M=0.16&E2&\\
SD=1.19 & E4 & &SD=0.37&E4& ***\\
\hline
E4 (Our counterfactual  & E1 & * & E4 & E1&***\\
explanation) \bf M=3.72 & E2 & & \bf M=0.24 & E2 & *** \\
SD=0.97 & E3 & &SD=0.42&E3&***\\
\hline
\end{tabular}
\end{table}

\section{Discussion And Conclusion}

We provided  a BlackBox \emph{Counterfactual Explainer}, designed to explain image classification models for medical applications.
 Our framework explains the decision by gradually transforming the input image to its counterfactual, such that the classifier's prediction is flipped. We have formulated and evaluated our framework on three properties of a valid counterfactual transformation: data consistency, classifier consistency, and self-consistency. Our results in Section~\ref{Desiderata} showed that our framework adheres to all three properties. 
 
 \emph{Comparison with xGEM and cycleGAN: } Our model creates visually appealing explanations that produce a desired outcome from the classification model while retaining maximum patient-specific information.  In comparison, both xGEM and cycleGAN failed on at least one essential property. xGEM model fails to create realistic images with a high FID score. Furthermore, the cycleGAN model fails to flip the classifier's decision with a low CV score $(\sim50\%)$.

Further, we present a thorough comparison between cycleGAN and our explanation in a human evaluation study.  The clinical experts' expressed high agreement that explanation images from cycleGAN were of high quality and they resembles the query CXR. But at the same time, users found the explanation images to be too similar to query CXR, and the cycleGAN explanations failed to provide the counterfactual reasoning for the decision.

In comparison, our explanation were most helpful in understanding the classification decision. Though the users reported inconsistencies in the visual appearance, but the overall sentiment looks positive and they selected our method as their preferred explanation method for improved understandability.

\emph{Clinical relevance of the explanations: } From a clinical perspective, we demonstrated that the counterfactual changes associated with normal (negative) or abnormal (positive) classification decisions are also associated with corresponding changes in disease-specific metrics such as CTR and SCP. In our clinical study, multiple radiologist reported using CTR as the metric to diagnose cardiomegaly. As radiologist annotations are expensive, and it is not efficient to perform human evaluation on a large test set, our results with CTR calculations provides a quantitative way t evaluate difference in real and counterfactual populations.

We acknowledge that our GAN-generated counterfactual explanations may have missing details such as small wires. In our extended experiments, we found that the foreign objects such as pacemaker have minimal importance in the classification decision (\emph{see} SM-Sec.~\ref{SM-ASPM}). We attempted to improve the preservation of such information through our revised context-aware reconstruction loss (CARL). However, even with CARL, the FO preservation score is not perfect.  A possible reason for this gap is the limited capacity of the object detector used to calculate the FOP score. Training a highly accurate FO detector is outside the scope of this study.

Further, a resolution of $256 \times 256$ for counterfactually generated images is smaller than a standard CXR. Small resolution limits the evaluation for fine details by both the algorithm and the interpreter. Our formulation of cGAN uses conditional-batch normalization (cBN) to encapsulate condition information while generating images. For efficient cBN, the mini-batches should be class-balanced. To accommodate high-resolution images with smaller batch sizes, we must decrease the number of conditions to ensure class-balanced batches. Fewer conditions resulted in a coarse transformation with abrupt changes across explanation images. In our experiments, we selected the largest $N$, which created a class-balanced batch that fits in GPU memory and resulted in stable cGAN training.  However, with the advent of larger-memory GPUs, we intend to apply our methods to higher resolution images in future work; and assess how that impacts interpretation by clinicians.

To conclude, this study uses counterfactual explanations as a way to audit a given black-box classifier and evaluate whether the radio-graphic features used by that classifier have any clinical relevance. In particular, the proposed model did well in explaining the decision for cardiomegaly and pleural effusions and was corroborated by an experienced radiology resident physician.
By providing visual explanations for deep learning decisions, radiologists better understand the causes of its decision-making. This is essential to lessen physicians' concerns regarding the ``BlackBox" nature by an algorithm and build needed trust for incorporation into everyday clinical workflow.  As an increasing amount of artificial intelligence algorithms offer the promise of everyday utility, counterfactually generated images are a promising conduit to building trust among diagnostic radiologists.

By providing counterfactual explanations, our work opens up many ideas for future work. Our framework showed that valid counterfactuals can be learned using an adversarial generative process that is regularized by the classification model. However, counterfactual reasoning is incomplete without a causal structure and explicitly modeling of the interventions. An interesting next step should explore incorporating or discovering plausible causal structures and creating explanations grounded with them.

\section*{Acknowledgments}
This work was partially supported by NIH Award Number 1R01HL141813-01, NSF 1839332 Tripod+X, SAP SE, and Pennsylvania’s Department of Health. We are grateful for the computational resources provided by Pittsburgh SuperComputing grant number TG-ASC170024.

\bibliographystyle{model2-names.bst}\biboptions{authoryear}
\bibliography{refs}

\section{Supplementary Material}

\subsection{Human evaluation}
\label{sm-he}
In our human evaluation study, we asked the following 15 questions for each CXR:
\begin{enumerate}
    \item Please provide your diagnosis for Cardiomegaly. Answers: Negative, mild, positive, not sure.
    \item (Only assessment) Do you agree with the AI system assessment for Cardiomegaly? Answers: yes, no
    \item (Only assessment)  I understand how the AI system made the above assessment for Cardiomegaly. Answers: 5-point Likert scale.
    \item (Assessment + SM) The heat-map is highlighting $<$blank$>$ important/relevant regions for Cardiomegaly. Answers: all, most, some, a few, none.
    \item (Assessment + SM) I understand how the AI system made the above assessment for Cardiomegaly. Answers: 5-point Likert scale.
    \item (Assessment + cycleGAN) The changes in the video are related to Cardiomegaly. Answers: 5-point Likert scale.
    \item (Assessment + cycleGAN)  I understand how the AI system made the above assessment for Cardiomegaly. Answers: 5-point Likert scale.
    \item (Assessment + cycleGAN) Images in the video look like a chest x-ray. Answers: 5-point Likert scale.
    \item (Assessment + cycleGAN) The images in the video look like the chest x-ray from the subject. Answers: 5-point Likert scale.
    \item (Assessment  + ours) The changes in the video are related to Cardiomegaly. Answers: 5-point Likert scale.
    \item (Assessment  + ours) Changes in the anatomy in the highlighted regions in the heat-map will change the assessment
of Cardiomegaly. Answers: 5-point Likert scale.
\item (Assessment + ours)  I understand how the AI system made the above assessment for Cardiomegaly. Answers: 5-point Likert scale.
\item (Assessment + ours) Images in the video look like a chest x-ray. Answers: 5-point Likert scale.
\item (Assessment + ours) The images in the video look like the chest x-ray from the subject. Answers: 5-point Likert scale.
\item Which explanation helped you the most in understanding the assessment made by the AI system
    Answers: Explanation-1: Heat-map highlighting important regions for assessment, Explanation-2: A video showing the transformation from negative to positive decision, Explanation-3: Two images at the two extreme ends of the decision (positive and negative), none.
\end{enumerate}

Next, we present the UI for different questions,
\begin{figure}[h!]
    \centering
    \includegraphics[width = 0.7\linewidth]
    {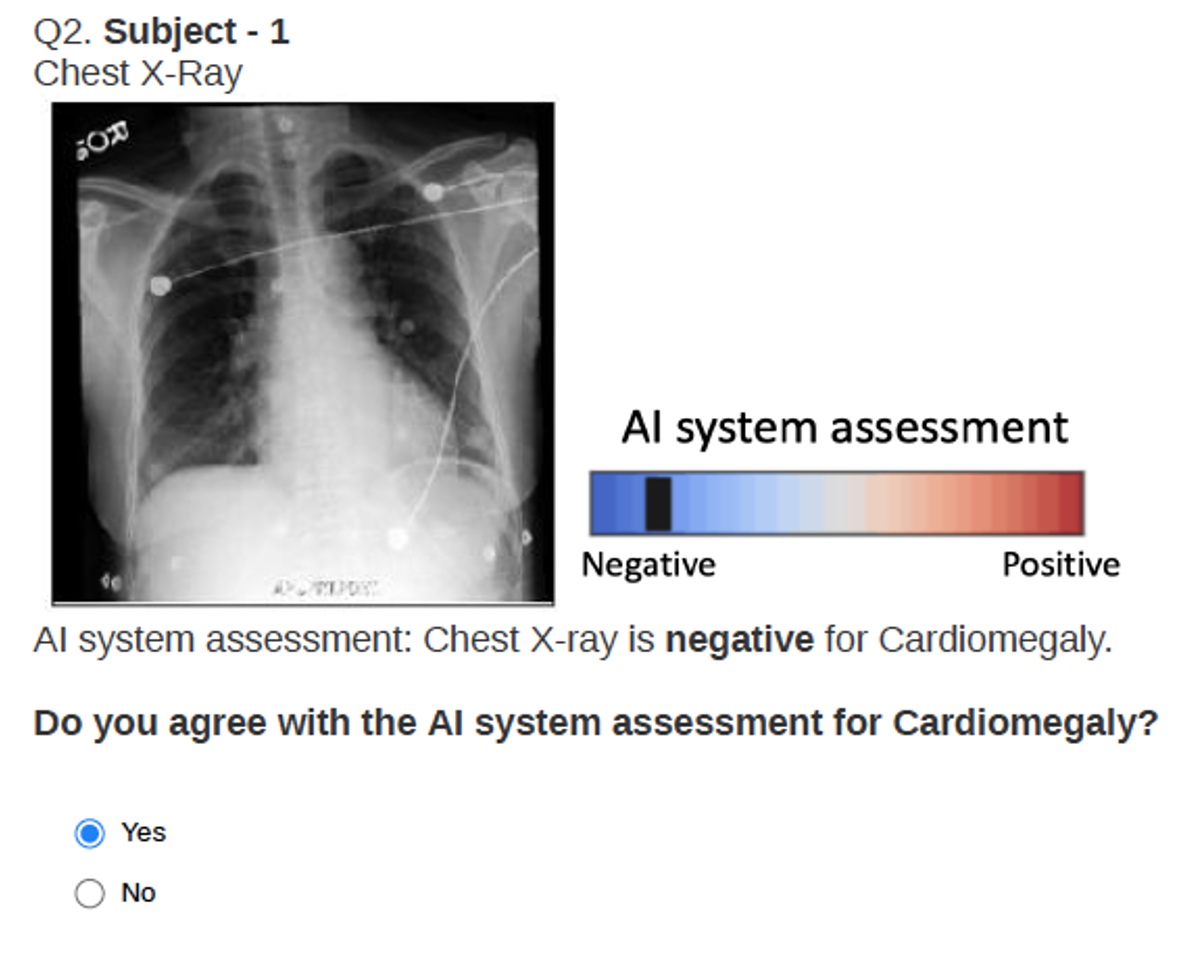}
    \caption{Question 2-3 showing the query CXR and the classifier's decision.}
\end{figure}

\begin{figure}[h!]
    \centering
    \includegraphics[width = 0.98\linewidth]
    {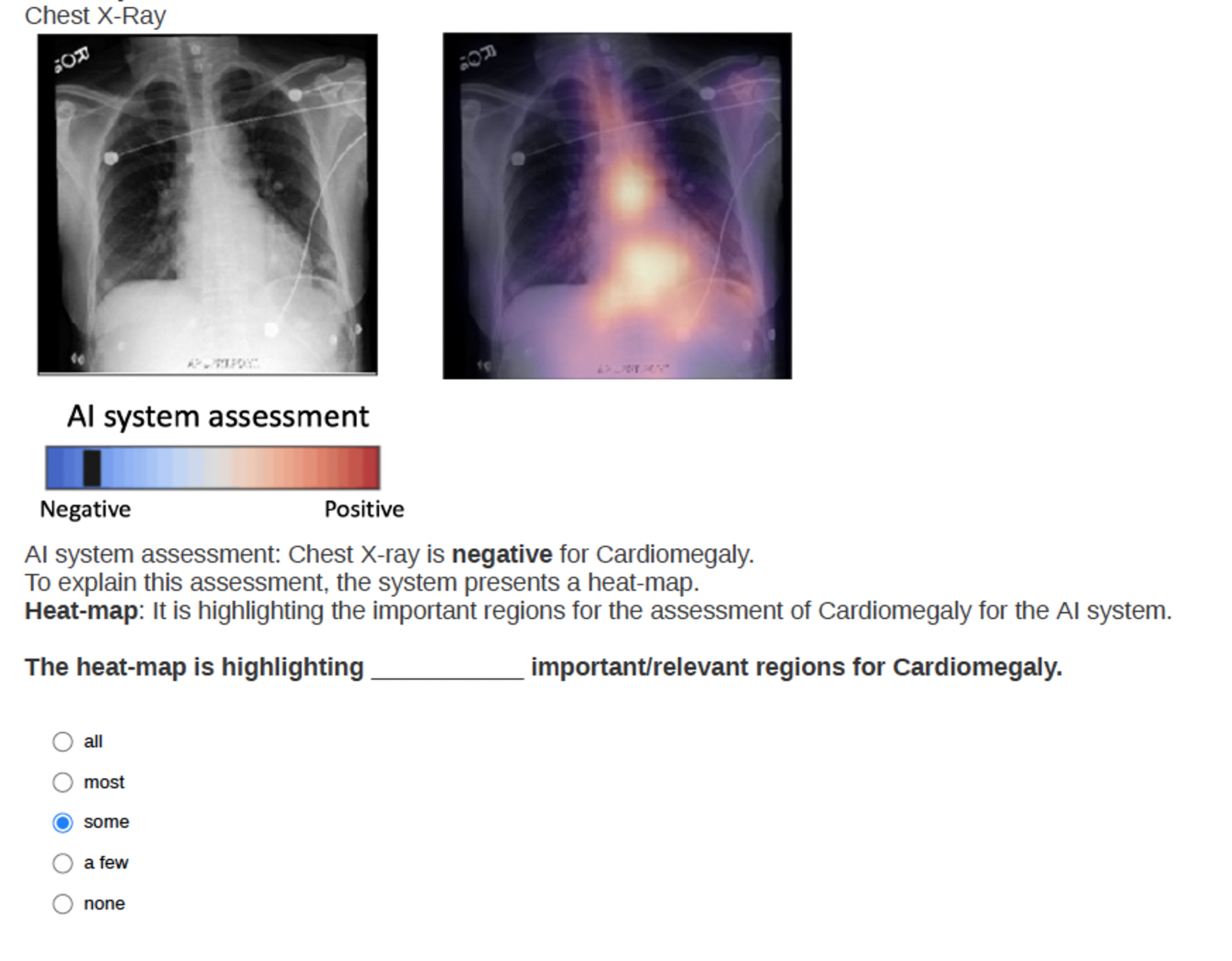}
    \caption{Question 4-5 showing the query CXR, the classifier's decision and the saliency map explanation.}
\end{figure}

\begin{figure}[h!]
    \centering
    \includegraphics[width = 0.98\linewidth]
    {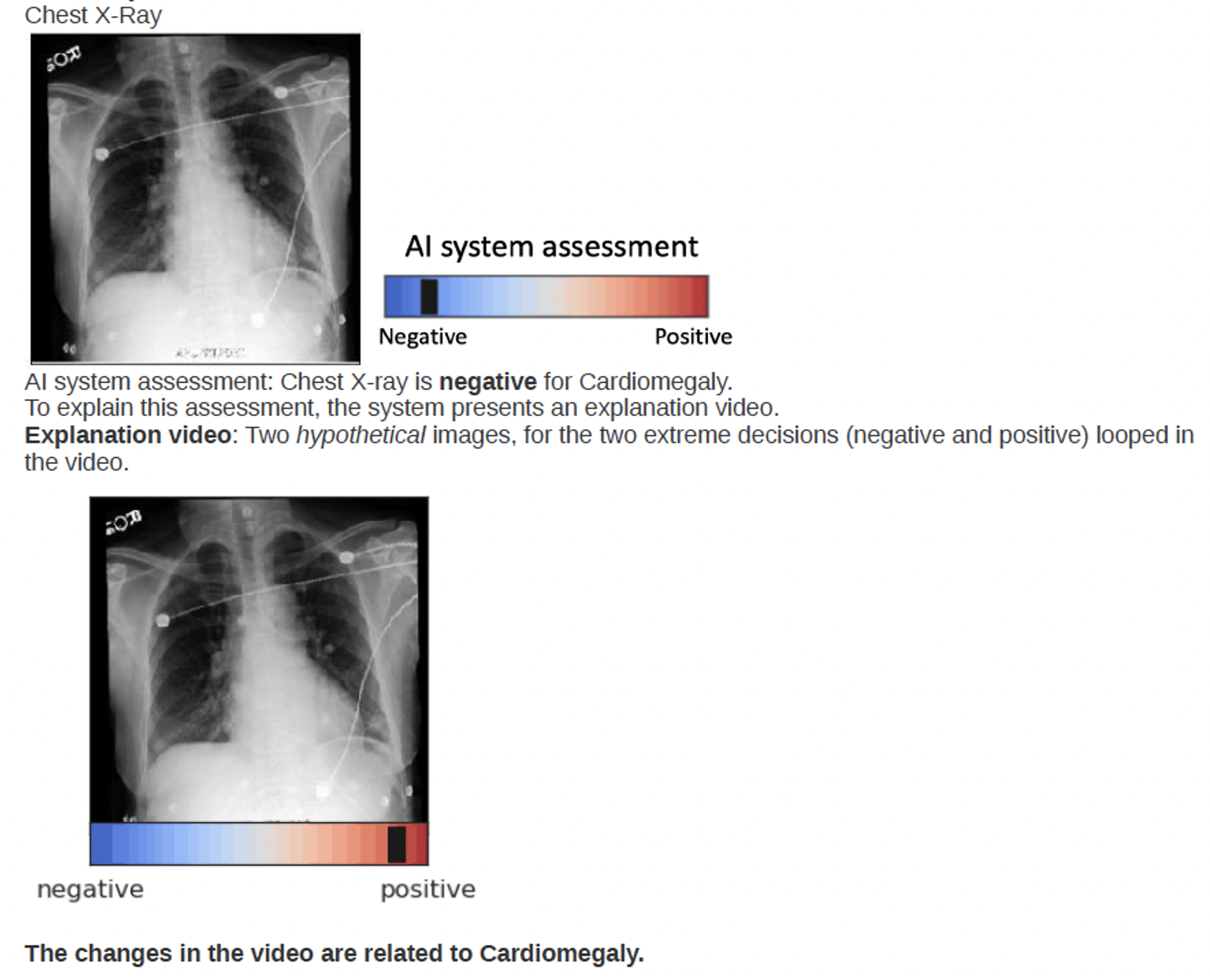}
    \caption{Question 6-9 showing the query CXR, the classifier's decision and the cycleGAN explanation.}
\end{figure}

\begin{figure}[h!]
    \centering
    \includegraphics[width = 0.98\linewidth]
    {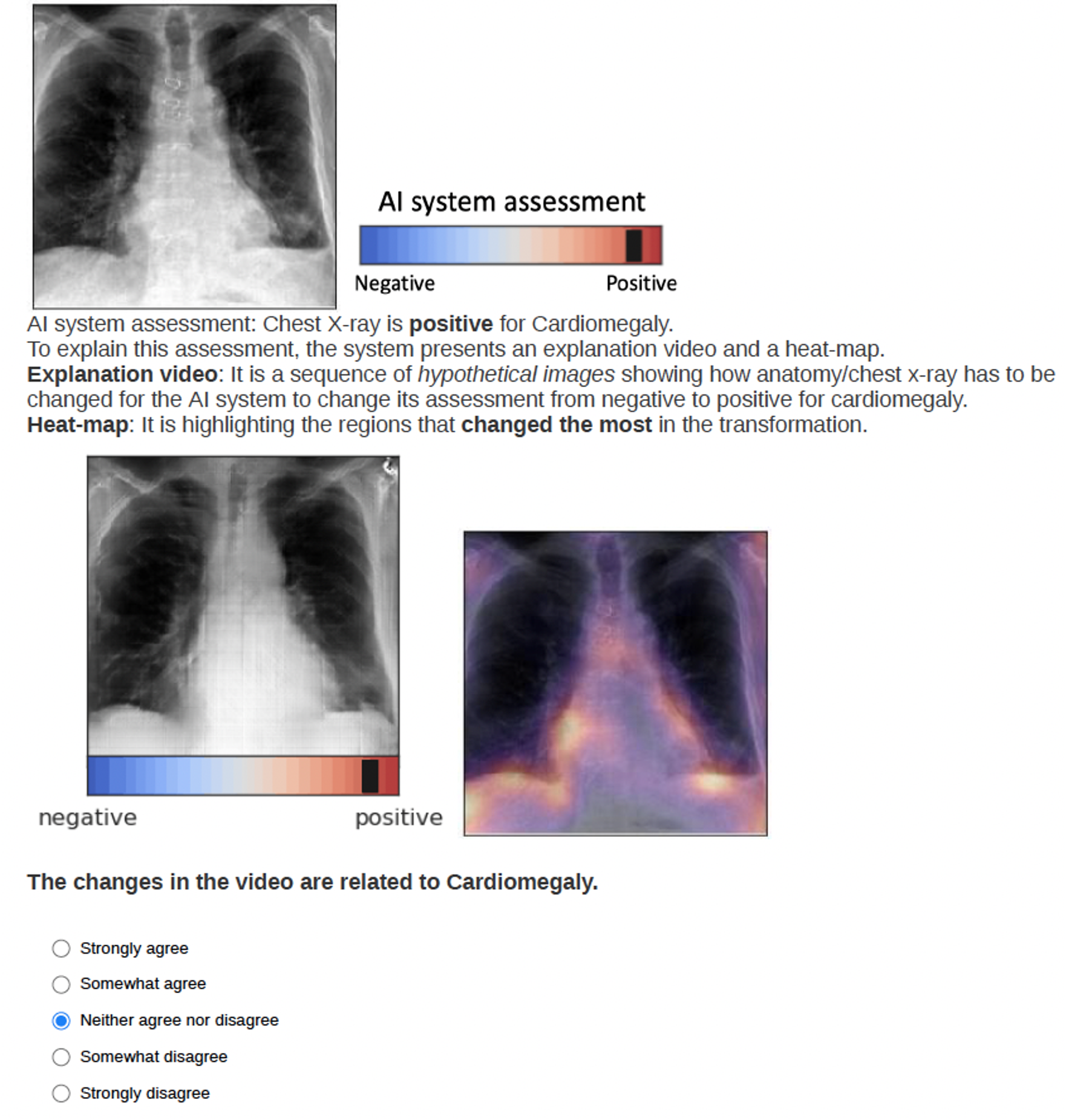}
    \caption{Question 10-14 showing the query CXR, the classifier's decision and our counterfactual explanation.}
\end{figure}

\subsection{Summarizing the notation}

\begin{table*}[h]
\centering
\caption{ Summarizing the notation}
\label{Notation-table}
\footnotesize
\begin{tabular}{l|l}
Notation &  Description\\
\hline
$\mathcal{X}$ & Input image space\\
$\rvx \in \mathcal{X}$ & Input image\\
$f:\mathcal{X}\rightarrow \mathcal{Y}$ & Pre-trained classification function\\
$f(\rvx)[k] \in [0,1]$ & Classifier's output for $k^{\text{th}}$ class \\
$\rvc$ & The condition used in cGAN, the desired classifier's output for $k^{\text{th}}$ class\\
$\rvx_{\rvc}$ & Explanation image\\
$f(\rvx_{\rvc})$ & Classifier's output for the explanation image\\
$\mathcal{I}_f(\rvx, \rvc)$ & Explanation function\\
$E(\cdot)$ & Image encoder\\
$\mathbf{z}$ & Latent representation of the input image\\
$C(\rvc)$&Discretizing function that maps $\rvc$ to an integer\\
$G(\mathbf{z}, \mathbf{c})$ & Generator of cGAN\\
$D(\rvx, \mathbf{c})$ & Discriminator of cGAN\\
$p_{\textnormal{data}}(\rvx)$ & Real image data distribution\\
$q(\rvx)$ & Learned data distribution by cGAN\\
$r(\rvx)$ & Loss term of cGAN that measures similarity between real and learned data distribution \\
$r(\mathbf{c}|\rvx)$ & Loss term of cGAN that evaluates correspondence between generated images and condition\\
$\phi(\rvx)$& Image feature extractor; part of the discriminator function\\
\hline
\end{tabular}
\end{table*}

Table.~\ref{Notation-table} summarizes the notation used in the manuscript.

\subsection{Dataset}
We focus on explaining classification models based on deep convolution neural networks (CNN); most state-of-the-art performance models fall in this regime. We used large, publicly available datasets of chest x-ray (CXR) images, MIMIC-CXR~\citep{Johnson2019MIMIC-CXRReports}. MIMIC-CXR dataset is a multi-modal dataset consisting of 473K CXR, and 206K reports from 63K patients.  We considered only frontal (posteroanterior PA or anteroposterior AP) view CXR. The datasets provide image-level labels for fourteen radio-graphic observations.  These labels are extracted from the radiology reports associated with the x-ray exams using an automated tool called the Stanford CheXpert labeler~\citep{Irvin2019CheXpert:Comparison}. The labeller first defines some thoracic observations using a radiology lexicon~\citep{Hansell2008FleischnerImaging}. It extracts and classifies (positive, negative, or uncertain mentions) these observations by processing their context in the report. Finally, it aggregates these observations into fourteen labels for each x-ray exam. For the MIMIC-CXR dataset, we extracted the labels ourselves, as we have access to the reports. 

\subsection{Classification Model}
\label{SM-CM}
To train the classifier, we considered the uncertain mention as a positive mention. We crop the original images to have the same height and width, then downsample them to 256 $\times$ 256 pixels. The intensities were normalized to have values between 0 and 1.  Following the approach in prior work~\citep{Rajpurkar2017CheXNet:Learning,Rubin2018LargeNetworks,Irvin2019CheXpert:Comparison} on diagnosis classification, we used DenseNet-121~\citep{Huang2016DenselyNetworks} architecture as the classification model.  In DenseNet, each layer implements a non-linear transformation based on composite functions such as Batch Normalization (BN), rectified linear unit (ReLU), pooling, or convolution. The resulting feature map at each layer is used as input for all the subsequent layers, leading to a highly convoluted multi-level multi-layer non-linear convolutional neural network.  We aim to explain such a model post-hoc without accessing the parameters learned by any layer or knowing the architectural details. Our proposed approach can be used for explaining any DL based neural network.

\begin{table}
\centering
\caption{ Explanation Model (cGAN) Architecture}
\label{Tab-arch}
(a) Encoder  \\  
\begin{tabular}{ c } 
\hline
\hline
Grayscale image $\rvx \in \mathbb{R}^{256 \times 256 \times 1}$ \\
BN, ReLU, 3$\times$3 conv 64 \\
Encoder-ResBlock down 128 \\
Encoder-ResBlock down 256 \\
Encoder-ResBlock down 512 \\
Encoder-ResBlock down 1024 \\
Encoder-ResBlock down 1024 \\
\\
(b) Generator\\
\hline
\hline
Latent code $\rvz \in \mathbb{R}^{1024}$\\
Generator-ResBlock up 1024, $\rvy$\\
Generator-ResBlock up 512, $\rvy$\\
Generator-ResBlock up 256, $\rvy$\\
Generator-ResBlock up 128, $\rvy$\\
Generator-ResBlock up 64, $\rvy$\\
BN, ReLU, 3$\times$3 conv 1\\
Tanh\\
\\
(c) Discriminator \\
\hline
\hline
Grayscale image $\rvx \in \mathbb{R}^{256 \times 256 \times 1}$ \\
Discriminator-ResBlock down 64\\
Discriminator-ResBlock down 128\\
Discriminator-ResBlock down 256\\
Discriminator-ResBlock down 512\\
Discriminator-ResBlock down 1024\\
Discriminator-ResBlock  1024\\
ReLU, Global Sum Pooling (GSP) $|$ Embed($\rvy$)\\
Inner Product (GSP, Embed($\rvy$)) $\rightarrow \mathbb{R}^1$ \\
Add(SN-Dense(GSP) $\rightarrow \mathbb{R}^1$, Inner Product)\\
\end{tabular} 
\end{table}

\subsection{Explanation Function}
\label{SM-EF}
\begin{figure*}[ht]
    \centering
    \includegraphics[width = 0.8\linewidth]
    {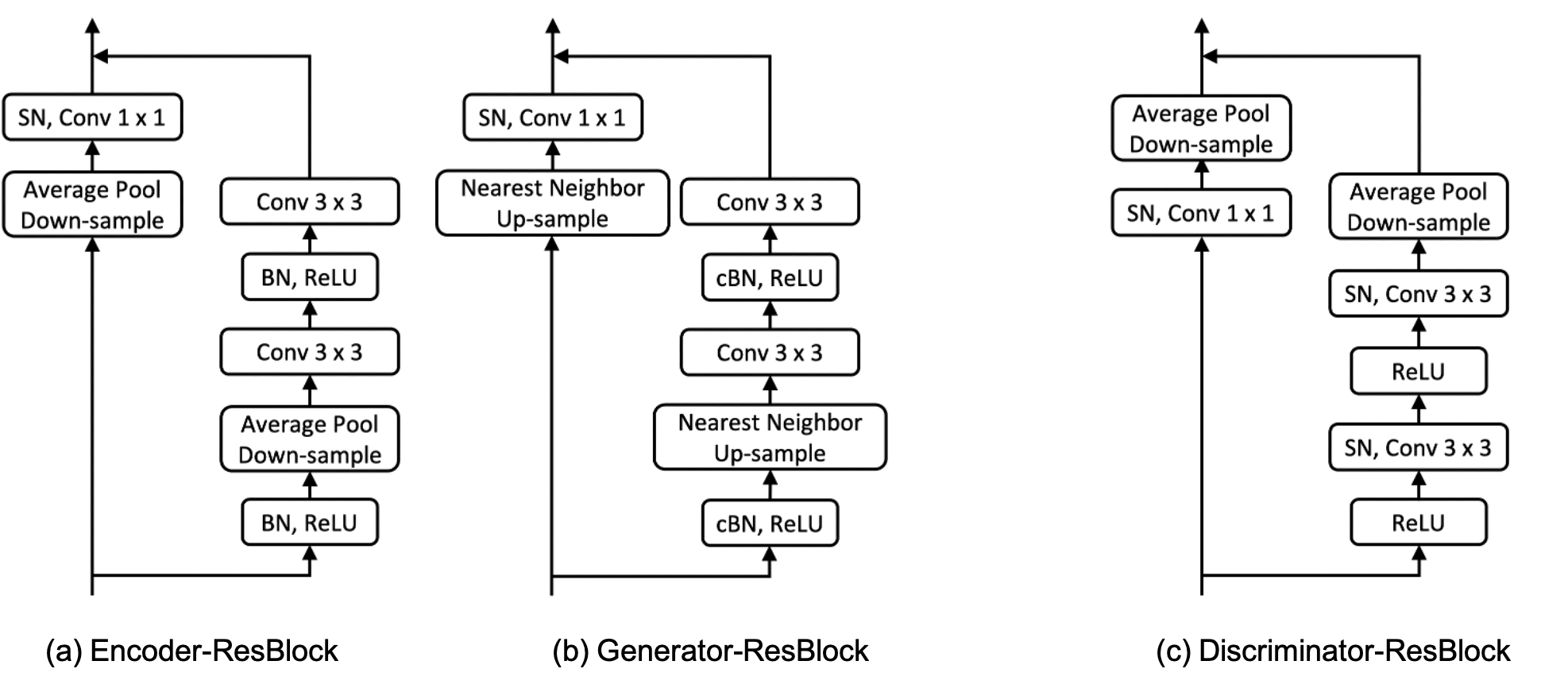}
    \caption{  Architecture of the ResBlocks used in all experiments.
    }
    \label{Fig_resblock}
\end{figure*}

The explanation function is a conditional GAN with an encoder. We used a ResNet~\citep{He2016DeepRecognition} architecture for the Encoder, Generator, and Discriminator. The details of the architecture are given in Table~\ref{Tab-arch}.  For the encoder network, we used five ResBlocks with the standard batch normalization layer (BN). In encoder-ResBlock, we performed down-sampling (average pool) before the first \textit{conv} of the ResBlock as shown in Fig.~\ref{Fig_resblock}.a. For the  generator network, we follow the details in~\citep{Miyato2018SpectralNetworks} and replace the BN layer in encoder-ResBlock with  conditional BN (cBN) to encode the condition (\textit{see} Fig.~\ref{Fig_resblock}.b.). The architecture for the generator has five ResBlocks; each ResBlock performed up-sampling through the nearest neighbour interpolator.  For the discriminator, we used spectral normalization (SN)~\citep{Miyato2018CGANsDiscriminator} in Discriminator-ResBlock and performed down-sampling after the second \textit{conv} of the ResBlock as shown in Fig.~\ref{Fig_resblock}.c. For the optimization, we used Adam optimizer~\citep{Kingma2014Adam:Optimization}, with hyper-parameters set to $\alpha = 0.0002, \beta_1 = 0, \beta_2 = 0.9$ and updated the discriminator five times per one update of the generator and encoder.

For creating the training dataset, we  divide the posterior distribution  for the target class, $f(\rvx) \in [0,1]$ into $N$ equally-sized bins. The cGAN is then trained on $N$ conditions.  For efficient training, cBN requires class-balanced batches. A smaller value for $\delta$ results in more conditions for training cGAN, increasing cGAN complexity and training time. Also,  we have to increase the batch size to ensure each condition is well represented in a batch. Hence, the GPU memory size bounds the high value for $N$. A small  $N$ is equivalent to fewer conditions, resulting in a coarse transformation which leads to abrupt changes across explanation images. In our experiments, we used $N=10$,  with a batch size of 32. We experimented with different values of $N$ and selected the largest $N$, which created a class-balanced batch that fits in GPU memory and resulted in stable cGAN training.

\subsection{Semantic Segmentation}
\label{SM-SS}
We adopted a 2D U-Net~\citep{Ronneberger2015U-NetSegmentation} to perform semantic segmentation, to mark the lung and the heart contour in a CXR. The network optimizes a multi-categorical cross-entropy loss function, defined as,
\begin{equation}\label{eq:ss}
\mathcal{L_{\theta}} := \sum_s \sum_i \mathbbm{1}(y_i = s) \log(p_{\theta}(x_i)) ,
\end{equation}
where $\mathbbm{1}$ is the indicator function, $y_i$ is the ground truth label for i-th pixel. $s$ is the segmentation label with values (background, the lung or the heart). $p_{\theta}(x_i)$ denotes the output probability for pixel $x_i$ and $\theta$ are the learned parameters. The network is trained on 385 CXRs and corresponding masks from Japanese Society of Radiological Technology (JSRT) ~\citep{vanGinneken2006SegmentationDatabase} and Montgomery~\citep{Jaeger2014TwoDiseases.} datasets.

\subsection{Object Detection}
\label{SM-OD}
We trained an object detector network to identify medical devices in a CXR. For the MIMIC-CXR dataset, we pre-processed the reports to extract keywords/observations that correspond to medical devices, including pacemakers, screws, and other hardware.  Such foreign objects are easy to identify in a CXR and do not requires expert knowledge for manual labelling.  Using the CheXpert labeller, we extracted 300 CXR images with positive mentions for each observation.  The extracted x-rays are then manually annotated with bounding box annotations marking the presence of foreign objects using the LabelMe~\citep{Wada2016LabelmePython} annotation tool. Next, we trained an object detector based on fFast Region-based CNN ~\citep{Ren2015FasterNetworks}, which used VGG-16 model~\citep{Simonyan2014VeryRecognition}, trained on the MIMIC-CXR dataset as its foundation. We used this object detector to enforce our novel context-aware reconstruction loss (CARL). 

\begin{figure}[h]
    \centering
    \includegraphics[width = 0.4\linewidth]
    {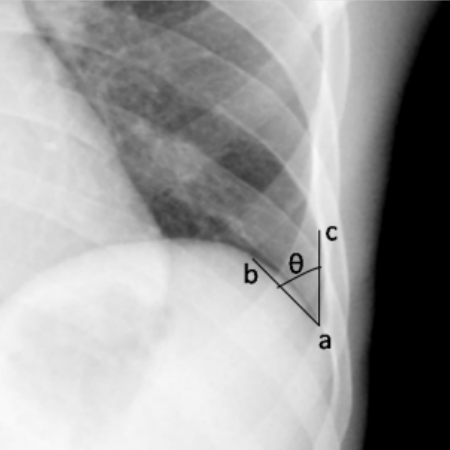}
    \caption{The costophrenic angle (CPA) on a CXR is marked as the angle formed by, (a) costophrenic angle point, (b) hemidiaphragm point and (c) lateral chest wall point, as shown by Maduskar \etal in~\citep{Maduskar2016AutomaticRadiographs}}
    \label{FIG_CPA}
\end{figure}

We trained similar detectors for identifying normal and abnormal CP recess regions in a CXR. We associated an abnormal CP recess with the radiological finding of a blunt CP angle as identified by the positive mention for \textit{``blunting of costophrenic angle"} in the corresponding radiology report. For the normal-CP recess, we considered images with a positive mention for \emph{``lungs are clear"} in the reports. We extracted 300 CXR images with positive mention of respective terms for normal and abnormal CP recess to train the object detector.

Please note that the object detector for CP recess is only used for evaluation purposes, and they were not used during the training of the explanation function. In literature, the blunting of CPA is an indication of pleural effusion~\citep{Maduskar2013AutomatedRadiographs,Maduskar2016AutomaticRadiographs}. The angle between the chest wall and the diaphragm arc is called the costophrenic angle (CPA). Marking the CPA angle on a CXR requires an expert to mark the three points, (a) costophrenic angle point, (b) hemidiaphragm point and (c) lateral chest wall point and then calculate the angle as shown in Fig.~\ref{FIG_CPA}.  Learning automatic marking of CPA angle requires expert annotation and is prone to error. Hence, rather than marking the CPA angle, we annotate the CP region with a bounding box which is a much simpler task. We then learned an object detector to identify normal or abnormal CP recess in a CXR and used the Score for detecting a normal CP recess (SCP) as our evaluation metric.

\subsection{xGEM}
\label{SM-xGEM}
We refer to work by Joshi \etal~\citep{Joshi2019TowardsSystems} for the implementation of xGEM. 
xGEM iteratively traverses the input image's latent space and optimizes the traversal to flip the classifier's decision to a different class.  Specifically, it solves the following optimization
\begin{equation}
    \Tilde{\rvx} = \mathcal{G}_{\theta}(\arg \min_{\mathbf{z}\in \mathbb{R}^d}\mathcal{L}(\rvx,\mathcal{G}_{\theta}(\mathbf{z})) + \lambda\ell(f(\mathcal{G}_{\theta}(\mathbf{z})),y^{'}))
\end{equation}
where the first terms is an $\ell_2$ distance loss for comparing real and generated data. The second term ensures that the classification decision for the generated sample is in favour of class $y^{'}$ and $y^{'} \neq y$ is a class other than original decision. Unless explicitly imposed, the explanation image does not look realistic. The explanation image is generated from an updated latent feature, and the expressiveness of the generator limits its visual quality.  xGEM adopted a variational autoencoder (VAE) as the generator. VAE uses a Gaussian likelihood ($\ell_2$ reconstruction), an unrealistic assumption for image data. Hence, vanilla VAE is known to produce over-smoothed images~\citep{Huang2018IntroVAE:Synthesis}.
 The VAE used is available at \href{https://github.com/LynnHo/VAE-Tensorflow}{https://github.com/LynnHo/VAE-Tensorflow}. All settings and architectures were set to default values. The original code generates an image of dimension 64x64. We extended the given network to produce an image with dimensions 256$\times$256.

\subsection{cycleGAN}
\label{SM-CG}
We refer to the work by Narayanaswamy \etal~\citep{Narayanaswamy2020ScientificTranslation} and DeGrave \etal~\citep{DeGrave2020AISignal} for the implementation details of cycleGAN. The network architecture for cycleGAN is replicated from the GitHub repository \href{https://github.com/junyanz/pytorch-CycleGAN-and-pix2pix}{https://github.com/junyanz/pytorch-CycleGAN-and-pix2pix}.

For training cycleGAN, we consider two sets of images. The first set comprises 2000 images from the MIMIC-CXR dataset such that the classifier has a  positive prediction for the presence of a target disease \ie $f(\rvx) > 0.9$, and the second set has the same number of images but with strong negative prediction \ie  $f(\rvx) < 0.1$. We train one such model for each target disease.

\subsection{Extended results for identity preservation}
A FO is critical in identifying the patient in an x-ray. FO's disappearance may lead to a false conclusion that removing FO resulted in the changed classification decision.

  \begin{figure}[h]
    \centering
    \includegraphics[width = 0.7\linewidth]
    {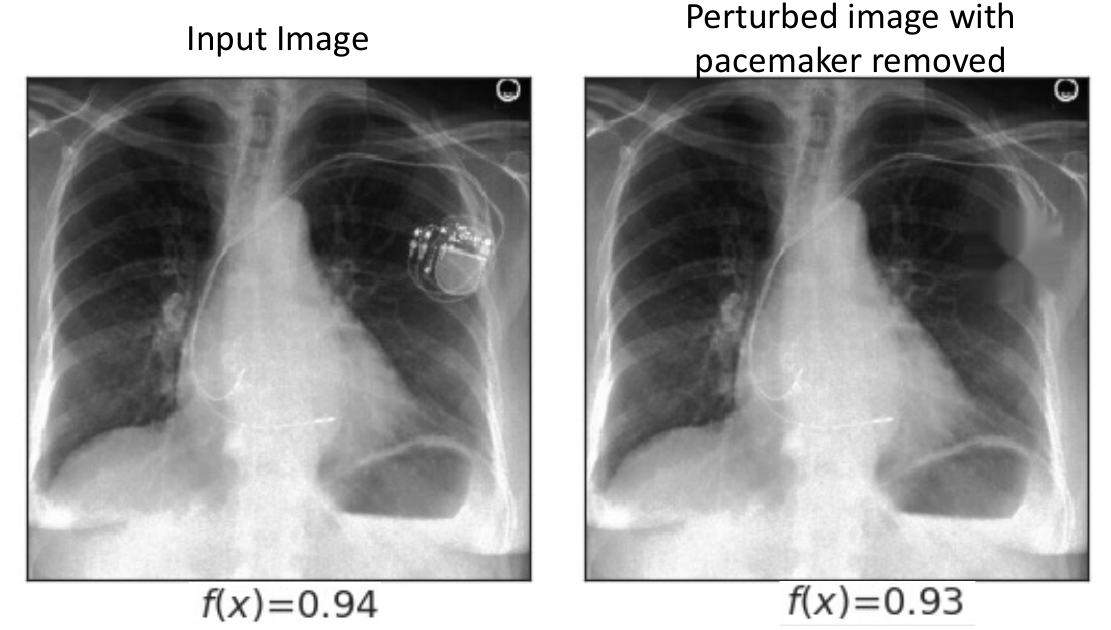}
    \caption{An example of input image before and after removing the pacemaker.}
    \label{Fig_pacemaker-1}
\end{figure}

\begin{figure*}[htbp]
    \centering
    \includegraphics[width = 0.7\linewidth]
    {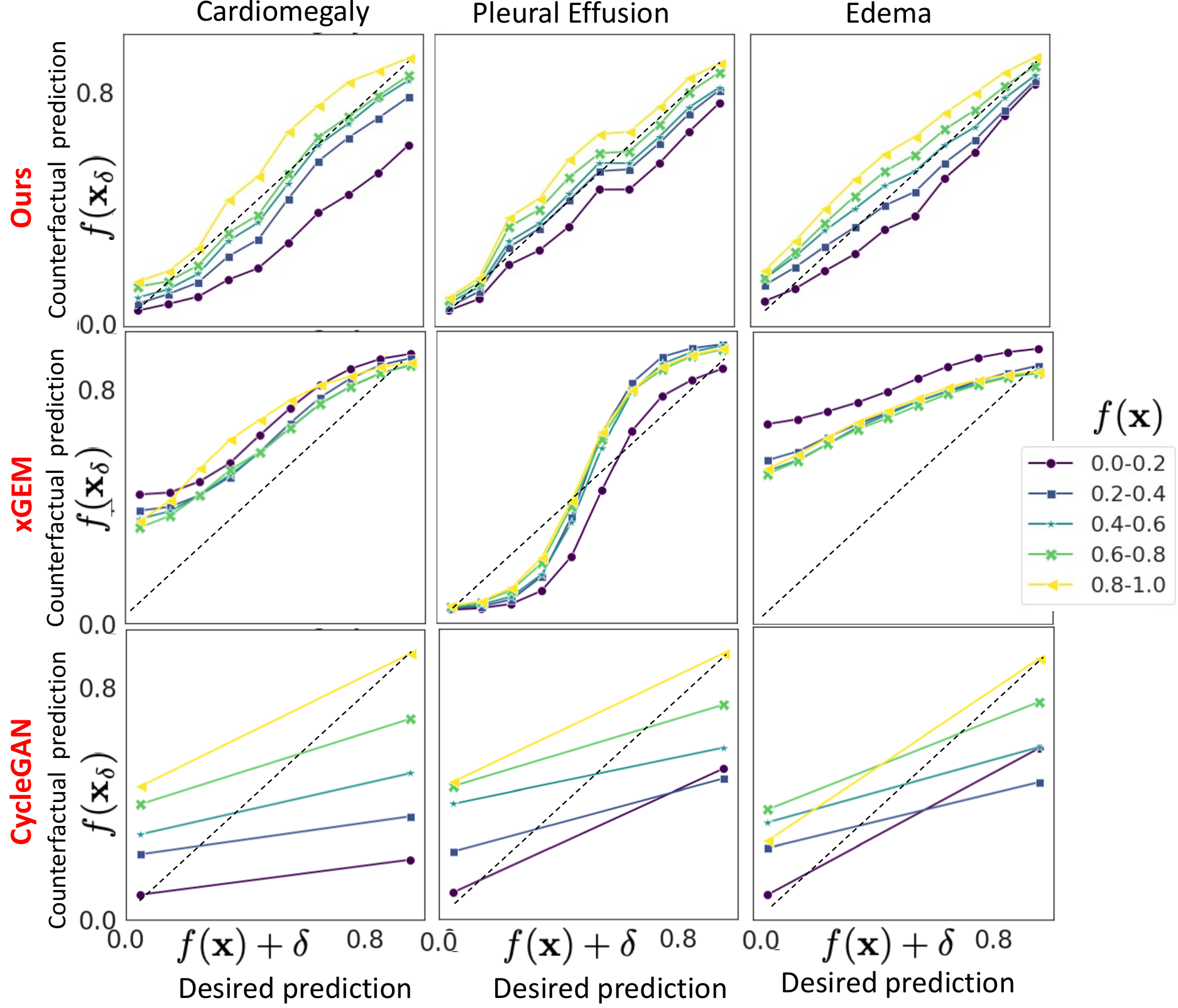}
    \caption{
    The plot of desired outcome, $f(\rvx) + \delta$, against actual response of the classifier
on generated explanations, $f(\rvx_{\delta})$. Each line represents a set of input images with classification prediction $f(\rvx)$ in a given range. Dashed line represents $y = x$ line.}
    \label{Fig_Cls_ext}
\end{figure*}

\subsubsection{Ablation study over pacemaker}
\label{SM-ASPM}
We performed an ablation study to investigate if a pacemaker is influencing the classifier's prediction for cardiomegaly. We consider 300 subjects that are positively predicted for cardiomegaly and have a pacemaker. We used our pre-trained object detector to find the bounding-box annotations for these images. Using the bounding-box, we created a perturbation of the input image by masking the pacemaker and in-filling the masked region with the surrounding context. 
    An example of the perturbation image is shown in Fig.~\ref{Fig_pacemaker-1}. We passed the perturbed image through the classifier and calculated the difference in the classifier's prediction before and after removing the pacemaker. The average change in prediction was negligible ($0.03$). Hence, pacemaker is not influencing classification decisions for cardiomegaly. 
  \begin{table}[h]
\centering
\caption{ The latent-space closeness (LSC) score for our model with and without the context-aware reconstruction loss (CARL).}
\label{OD-table1}
\begin{tabular}{c|c}
Foreign &  \bf LSC score\\
 Object &  CARL better than $\ell_1$  \\
\hline
Pacemaker & 0.79  \\
Hardware & 0.87    \\
\end{tabular}
\end{table}

\subsubsection{Latent space closeness (LCS)}
\begin{figure*}[ht]
    \centering
    \includegraphics[width = 0.9\linewidth]
    {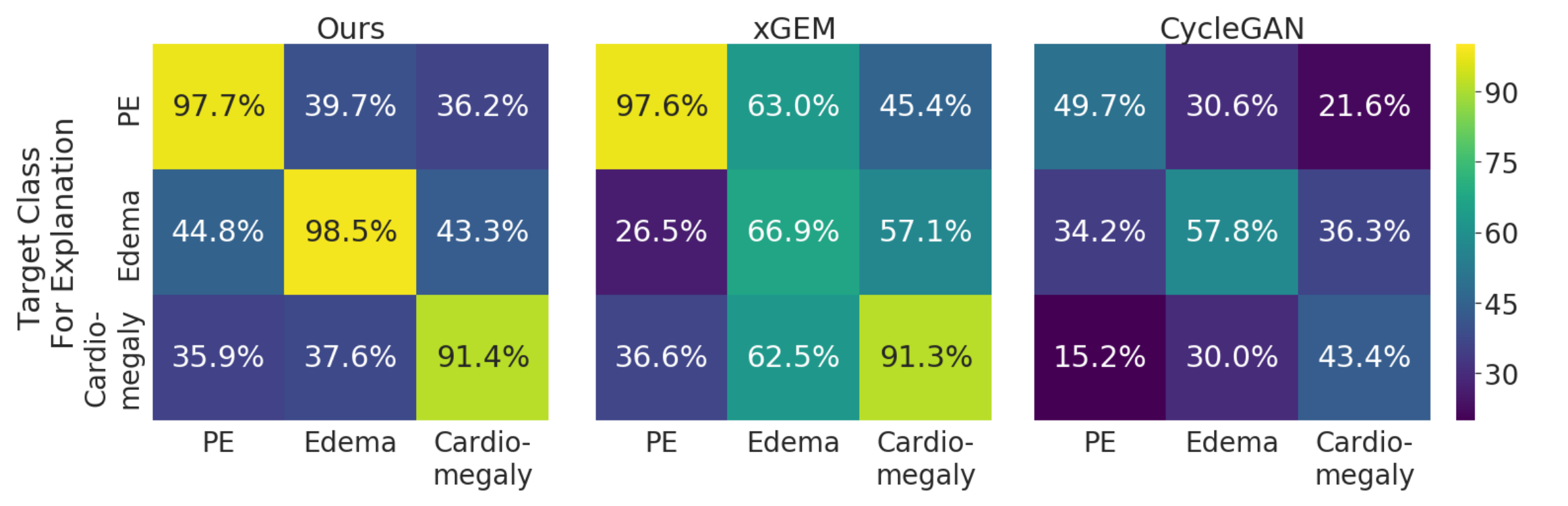}
    \caption{
     Each cell is the fraction of the generated explanations, that have flipped in a class as compared to the query image. The x-axis shows the classes in a multi-label setting, and the y-axis shows the target class for
which an explanation is generated. Note: This is not a confusion matrix. }
    \label{Fig_multi}
\end{figure*}

We compared the explanations generated using CARL against those generated using simple $\ell_1$ reconstruction loss on their similarity with the input images. To quantify the similarity between the explanation images and the query image in a latent space, we used latent-space closeness (LSC) score.  LSC score is the fraction of the images where explanation image  derived using CARL ($\rvx_{\rvc}^{\textnormal{CARL}}$) is closest to the query image $\rvx$ as compared to explanations generated using $\ell_1$ loss \ie $\rvx_{\rvc}^{\ell_1}$. We calculated similarity as the euclidean distance between the embedding for the query and explanation images. LSC score is defined as,
\[LSC = \sum_{\rvx \in \mathcal{X}, \rvc} \mathbbm{1} \left( \langle E(\rvx), E(\rvx_{\rvc}^{\textnormal{CARL}}) \rangle < \langle E(\rvx), E(\rvx_{\rvc}^{\ell_1}) \rangle \right)\]
 where $E(\cdot)$ is a pre-trained feature extractor based on the Inception v3 network. Table~\ref{OD-table1} presents our results. A high LSC score, together with a high CV score (Fig.~\ref{Fig_multi}) shows that the query and counterfactual images are fundamentally same but differs only in features that are sufficient to flip the classification decision.

\subsection{Extended classifier consistency results}

Our explanation framework gradually perturbs the input image to traverse the classification boundary from one extreme (negative) to another (positive). We quantify the consistency between our explanations and the classification model at every step of this transformation. We divided the prediction range $[0,1]$ into ten equally sized bins. For each bin, we generated an explanation image by choosing an appropriate, $\rvc \in [0,1]$. We further divided the input image space into five groups based on their initial prediction \ie $f(\rvx)$. In Fig.~\ref{Fig_Cls_ext}, we represented each group as a line and plotted the average response of the classifier \ie $f(\rvx_{\rvc})$ for explanations in each bin against the expected outcome \ie $\rvc$.  For xGEM, we generated multiple, progressively changing explanations by traversing the latent space. For each input image, we generated ten explanation images. For cycleGAN, we can generate only images at the two extreme ends of the decision boundary.

\begin{table*}[t]
\caption{ FID score quantifies the visual appearance of the explanations. CV score is the fraction of explanations that have an opposite prediction compared to the input image. FOP score is the fraction of real images with FO, in which FO was also detected in the corresponding explanation image. In configuration with $\lambda_1 = 0$ there is no adversarial loss from cGAN, in  $\lambda_2 = 0$ there is no KL -loss for classifier consistency and in $\lambda_3 = 0$ there is no context-aware self reconstruction loss.     }
\label{Ablation-table}
\small	
\begin{center}
\begin{tabular}{c|cccc|cccc|cccc}
\multicolumn{1}{c}{ }  &  \multicolumn{4}{c}{\bf Cardiomegaly} &  \multicolumn{4}{c}{\bf Pleural Effusion}  &  \multicolumn{4}{c}{\bf Edema}\\
& Baseline & $\lambda_1$ = 0 & $\lambda_2$ = 0 &$\lambda_3$ = 0  & Baseline & $\lambda_1$ = 0 & $\lambda_2$ = 0 &$\lambda_3$ = 0  &Baseline & $\lambda_1$ = 0 & $\lambda_2$ = 0 &$\lambda_3$ = 0  \\
\hline
\multicolumn{13}{c}{\bf FID score} \\
\hline
Normal&  166 & 200& 174 & 160 &146  & 210 & 150 & 149& 149 & 169 & 153 & 155\\
Abnormal & 137 & 189 & 138 & 140 & 122& 178 & 120 & 130& 102 & 170 & 109 & 120\\
\hline
\hline
\multicolumn{13}{c}{\bf Counterfactual Validity (CV) Score} \\
\hline
Real ($f(\rvx) \in [0,1]$)&  0.91&  0.89 & 0.43 & 0.92 &  0.97& 0.93 & 0.43 & 0.97 &  0.98 & 0.95 & 0.45 & 0.91\\
\hline
\multicolumn{13}{c}{\bf Foreign Object Preservation (FOP) score} \\
\hline
Pacemaker&  0.52&  0.2 & 0.55 & 0.19\\
\hline
\end{tabular}
\end{center}
\end{table*}

Fig.~\ref{Fig_Cls_ext} shows our results. It an extension of the results in Fig.~\ref{Fig_Cls}.  The positive slope of the line-plot, parallel to $y = x$ line  confirms that starting from images with low $f(\rvx)$, our model creates fake images such that $f(\rvx_{\rvc})$ is high and vice-versa.  Thus, our model creates explanations that successfully flips the classification decision and, hence, represents the decision-making process of the classifier. In contrast, for cycleGAN model, if $f(\rvx) \in [0.0,0.4]$ (blue line-plot), the resulting explanations have $f(\rvx_{\rvc}) < 0.5$, hence, cycleGAN model fails to flip the classification decision, as also evident in low CV score in Table.~\ref{FID-table}.

\subsection{Evaluating class discrimination}
In multi-label settings, multiple labels can be true for a given image. A multi-label setting is common in CXR diagnosis. For example, cardiomegaly and pleural effusion are associated with cardiogenic edema and frequently co-occur in a CXR. Please note that our classification model is also trained in a multi-label setting where the fourteen radiological findings may co-occur in a CXR. In this evaluation, we demonstrate the sensitivity of our generated explanations to the task being explained. We considered three diagnosis tasks, cardiomegaly, pleural effusion, and edema. For each task, we trained one explanation model. Ideally, an explanation model trained to explain a given task should produce explanations consistent with the query image on all the other classes besides the given task. Fig.~\ref{Fig_multi} plots the fraction of the generated explanations, that have flipped in other classes as compared to the query image. Ideally, the fraction should be maximum for the given task and small for the rest of the classes. In Fig.~\ref{Fig_multi}, each column represents one task, and each row is one run of our method to explain a given task. The diagonal values also represent the counterfactual validity (CV) score reported in Table.~\ref{FID-table}.

\begin{figure*}[h]
    \centering
    \includegraphics[width = 0.9\linewidth]
    {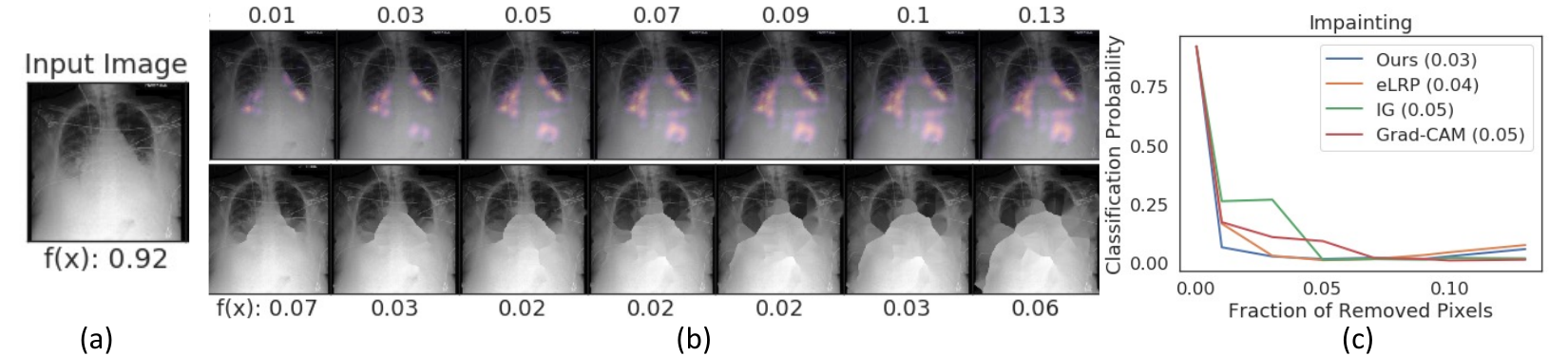}
    \caption{\footnotesize Deletion-by-impainting: (a) input image. (b) transformation of the input image as important pixels are deleted, and the resulting patches are in-filled base on the surrounding context. The importance is derived from the saliency map produced from our (top-row) and gradient-based (bottom-row) method. The top label shows the fraction of removed pixels. The bottom label shows the classification outcome for a target class. (c) The plot shows the change in classification prediction as a function of the fraction of removed pixels. }
    \label{Fig_SM}
\end{figure*}

\subsection{Ablation Study}
\label{SM-AS}
Eq.~\ref{eq_final} shows our final loss function:
 We have three types of loss functions: adversarial loss from cGAN $\mathcal{L}_{\text{cGAN}}(D,G)$, KL loss $\gL_{f}(D,G)$, and CARL reconstruction loss $\gL_{\text{identity}}(E,G)$. The three losses enforce the three properties of our proposed explanation function: data consistency, classifier consistency, and context-aware self-consistency, respectively. In the ablation study, we quantify the importance of each of these components by training different models, where on hyper-parameter is set to zero while rest are equivalent ($\lambda_{1} = 1$, $\lambda_2 = 1$ and $\lambda_{3} = 0.5$). For \textbf{data consistency}, we evaluate Fr\'echet Inception Distance (FID). FID  score measures the visual quality of the generated explanations by comparing them with the real images.  We examined real and synthetic (\ie generated explanations) images on the two extreme of the decision boundary, \ie a normal group ($f(\rvx) < 0.2$) and an abnormal group  ($ f(\rvx) > 0.8$).  For \textbf{classifier consistency},  we reported results on counterfactual validity (CV) score. CV score is the fraction of counterfactual explanations that successfully flipped the classification decision i.e., if the input image is negative (normal) then the generated explanation is predicted as positive (abnormal) for the specific classification task. For \textbf{self consistency}, we calculated the FO preservation (FOP) score. FOP score is the fraction of real images, with successful detection of FO, in which FO was also detected in the corresponding explanation image $\rvx_{\delta}$.   
Table~\ref{Ablation-table} summarizes our results. In the absence of adversarial loss from cGAN ($\lambda_1 = 0$), FID score is very high and the FOP score is low as the generated images looks very different from the real CXR images.  When KL loss for classifier consistency is missing ($\lambda_2 = 0$), the CV score is poor as the generated explanations are derived without considering the classification function and hence they failed to flip the classification decision. In the absence of CARL loss ($\lambda_3 = 0$), the generated explanations are no longer for the same patient as in query CXR, hence FO in query CXR are absent in generated explanations, resulting in low FOP score.

\begin{table*}[h]
\caption{\small Results of independent t-test. We compared the difference distribution of cardiothoracic ratio (CTR) for cardiomegaly and the Score for normal Costophrenic recess (SCP) for pleural effusion.
}
\label{t-table}
\footnotesize
\centering
\begin{tabular}{c|cc|cccc|c|c|c}
Target  & & & \multicolumn{4}{c|}{\bf Paired Differences} &  & & \\
Disease & Real & Counterfactual& & & \multicolumn{2}{c|}{95\% Confidence Interval} & & & \\
& Group & Group & Mean Difference & Std & Lower & Upper & t & df & p-value \\
\hline
Cardiomegaly & $\mathcal{X}^n$ & $\mathcal{X}_{\rvc}^{n \rightarrow p}$ & \bf -0.03 & 0.07 & -0.03 & -0.01 & -4.4 & 304 & $<$ 0.0001 \\
(CTR) & $\mathcal{X}^p$ & $\mathcal{X}_{\rvc}^{p \rightarrow n}$ & \bf 0.14 & 0.12 & 0.13 & 0.15 & 24.7 & 513 & $\ll$ 0.0001 \\
Pleural effusion & $\mathcal{X}^n$ & $\mathcal{X}_{\rvc}^{n \rightarrow p}$ & \bf 0.13 & 0.22 & 0.06 & 0.13 & 5.9 & 217 & $\ll$ 0.0001 \\
(SCP)  & $\mathcal{X}^p$ & $\mathcal{X}_{\rvc}^{p \rightarrow p}$ & \bf -0.19 & 0.27 & -0.18 & -0.09 & -6.7 & 216 & $\ll$ 0.0001 \\
\hline
\hline
  & & & \multicolumn{4}{c|}{\bf Un-Paired Differences} &  & & \\
 &  &  & Mean & Mean & \multicolumn{2}{c|}{95\% Confidence Interval} & & & \\
& & & Real Group & Counterfactual Group & Lower & Upper & t & df & p-value \\
\hline
Cardiomegaly & $\mathcal{X}^n$ & $\mathcal{X}_{\rvc}^{p \rightarrow n}$ & \bf 0.46 & 0.42 & 0.02 & 0.06 & 5.2 & 817 & $<$ 0.0001 \\
(CTR) & $\mathcal{X}^p$ & $\mathcal{X}_{\rvc}^{n \rightarrow p}$ & \bf0.56 & 0.50 & 0.04 & 0.07 & 9.9 & 817 & $\ll$ 0.0001 \\
Pleural effusion & $\mathcal{X}^n$ & $\mathcal{X}_{\rvc}^{p \rightarrow n}$ & \bf0.69 & 0.61 & 0.18 & 0.27 & 9.3 & 433 & $\ll$ 0.0001 \\
(SCP)  & $\mathcal{X}^p$ & $\mathcal{X}_{\rvc}^{n \rightarrow p}$ & 0.42 & \bf0.56 & -0.32 & -0.21 & -9.7 & 433 & $\ll$ 0.0001 \\
\hline
\end{tabular}
\end{table*}
\begin{figure*}[htbp]
    \centering
    \includegraphics[width = 0.9\linewidth]
    {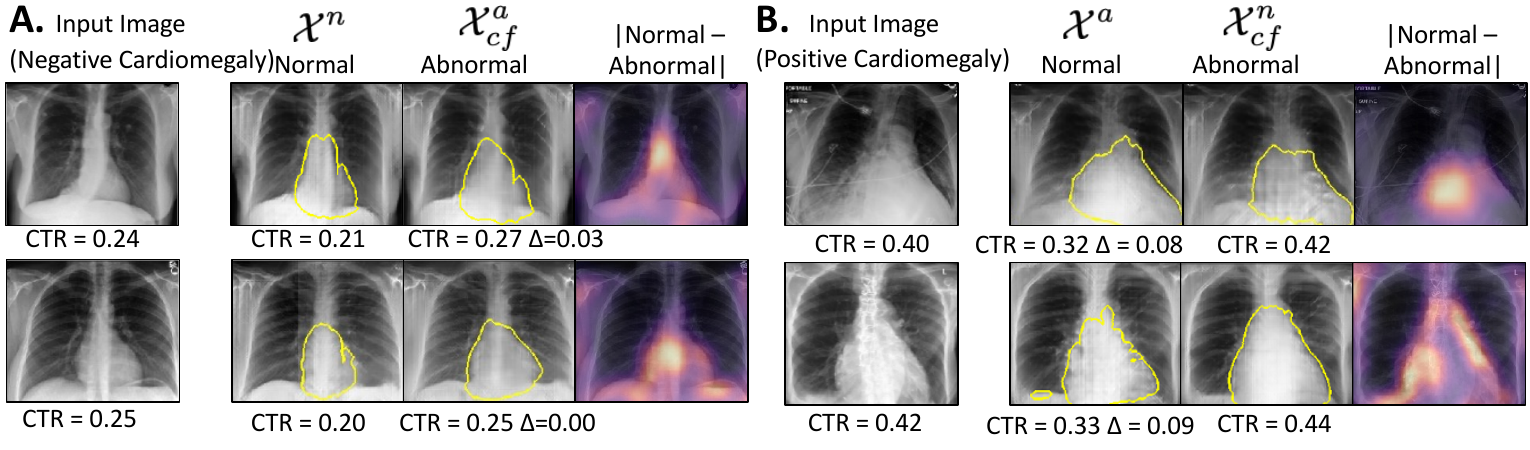}
    \caption{
    Extended results for explanation produced by our model for \textbf{Cardiomegaly}. For each image, we generate a normal and an abnormal explanation image. We show pixel-wise difference of the two generated images as the saliency map. In column A.(B.), we show input images negatively (positively) classified for Cardiomegaly. The yellow contour shows the heart boundary learned by a segmentation network. CTR is the cardiothoracic ratio. For column A, we observe a relatively minor change in CTR ($\Delta$) between real and counterfactual images than in column B. }
    \label{Fig_c}
\end{figure*}

\subsection{Extended results for saliency maps}

Our method doesn't produce a saliency map by default. We approximated  a saliency map as an absolute difference map between the explanations generated for the two extremes (normal with $f(\rvx_{\rvc}) < 0.2$ and abnormal $f(\rvx_{\rvc}) > 0.8$) of the decision function $f$. We show an example of saliency map generated by our method in Fig.~\ref{Fig_saliency}. Fig.~\ref{Fig_c} shows our extended results.

We also compared the saliency maps generated by our model with popular gradients based methods. For quantitative evaluation, we consider the \textit{deletion} evaluation metric~\citep{Petsiuk2018RISE:Models}. The metric quantifies how the probability of the target-class changes as important pixels are removed from an image. To remove pixels from an image, we tried selectively impainting the region based on its surroundings. In Fig.~\ref{Fig_SM}, we show an example of deletion-by-impainting. For generating results in Table.~\ref{table}, we plot the deletion curve for 500 images, and calculated area under the deletion curve (AUDC) for each.

Please note that, as more pixels are removed, the modified images become unrealistic and visually appear different from a CXR. The behavior of the classifier on such images is inconsistent. Low AUDC demonstrates that all the methods are successful in localizing the important regions for classification. However, unlike saliency-based methods, our counterfactual explanation provides extra information on \textit{what} image features in those relevant regions for classification and \textit{how} those image features should be modified to flip the decision.

\subsection{Disease-specific evaluation}

For quantitative analysis, 
we randomly sample two groups of real images (1) a \emph{real-normal} group defined as $\mathcal{X}^{n} = \{\rvx; f(\rvx) < 0.2\}$. It consists of real CXR images that are predicted as normal by the classifier $f$. (2) A \emph{real-abnormal} group defined as $\mathcal{X}^{p} = \{\rvx; f(\rvx) > 0.8\}$. For $\mathcal{X}^{n}$, we generated a counterfactual group as,  $\mathcal{X}_{\rvc}^{p} = \{\rvx \in \mathcal{X}^n; f(\mathcal{I}_f(\rvx, \rvc)) > 0.8\}$. Similarly for $\mathcal{X}^p$, we derived a counterfactual group as   $\mathcal{X}_{\rvc}^{n} = \{\rvx \in \mathcal{X}^p; f(\mathcal{I}_f(\rvx, \rvc)) < 0.2\}$.

Next, we quantify the differences in real and counterfactual groups by performing statistical tests on the distribution of clinical metrics such as cardiothoracic ratio (CTR) and the Score of normal Costophrenic recess (SCP). Specifically, we performed the dependent t-test statistics on clinical metrics for paired samples ($\mathcal{X}^n$ and $\mathcal{X}^p_{\rvc}$), ($\mathcal{X}^p$ and $\mathcal{X}^n_{\rvc}$) and the independent two-sample t-test statistics for normal ($\mathcal{X}^n$, $\mathcal{X}^n_{\rvc}$) and abnormal ($\mathcal{X}^p$, $\mathcal{X}^p_{\rvc}$) groups. The two-sample t-tests are statistical tests used to compare the means of two populations. A low p-value $< $ 0.0001 rejects the null hypothesis and supports the alternate hypothesis that the difference in the two groups is statistically significant and that this difference is unlikely to be caused by sampling error or by chance. For paired t-test, the mean difference corresponds to the average causal effect of the intervention on the variable under examination. In our setting, intervention is a \textit{do} operator on input image ($\rvx$), before intervention, resulting in a counterfactual image ($\rvx_{\rvc}$), after intervention.

Table~\ref{t-table} provides the extended results for the Fig.~\ref{Fig_Box}. Patients with cardiomegaly have higher CTR as compared to normal subjects. Hence, one should expect CTR($\mathcal{X}^n$) $<$ CTR($\mathcal{X}_{\rvc}^p$) and likewise CTR($\mathcal{X}^p$) $>$ CTR($\mathcal{X}_{\rvc}^n$). Consistent with clinical knowledge, in Table.~\ref{t-table}, we observe a negative mean difference of -0.03 for CTR($\mathcal{X}^n$) $-$ CTR($\mathcal{X}_{\rvc}^p$) (a p-value of $< 0.0001$) and a positive mean difference of 0.14 for CTR($\mathcal{X}^p$) $-$ CTR($\mathcal{X}_{\rvc}^n$) (with a p-value of $\ll 0.0001$). On  a  population-level  CTR  was  successful  in capturing  the  difference  between  normal  and  abnormal  CXRs.  Specifically in un-paired differences, we observe a low mean CTR values for normal subjects \ie mean CTR($\mathcal{X}^n$) = 0.46 as compared to mean CTR for abnormal patients \ie mean CTR($\mathcal{X}^p$) = 0.56. The low p-values supports the alternate hypothesis that the difference in the two groups is statistically significant.

Further, in Fig~\ref{Fig_c}.A, we show samples from input images that were predicted as negative for cardiomegaly ($\mathcal{X}^{n}$). In their counterfactual abnormal images (third column), we observe small changes in CTR are sufficient to flip the classification decision. This is consistent with a small mean difference CTR($\mathcal{X}^{n}$) - CTR($\mathcal{X}^{a}_{\rvc}) = -0.03$.
 In contrast, when we generate counterfactual normal (sixth column) from real abnormal images (positive for cardiomegaly, Fig~\ref{Fig_c}.B), significant changes in CTR lead to flipping of the prediction decision. This observation is consistent with a large mean difference CTR($\mathcal{X}^{a}$) - CTR($\mathcal{X}^{n}_{\rvc}) = 0.14$.

By design, the object detector assigns a low SCP to any indication of blunting CPA or abnormal CP recess. Hence, SCP($\mathcal{X}^n$) $>$ SCP($\mathcal{X}_{\rvc}^p$) and likewise SCP($\mathcal{X}^p$) $<$ SCP($\mathcal{X}_{\rvc}^n$). Consistent with our expectation, in Table.~\ref{t-table}, we observe a positive mean difference of 0.13 for SCP($\mathcal{X}^n$) $-$ SCP($\mathcal{X}_{\rvc}^p$) (with a p-value of $\ll 0.0001$) and a negative mean difference of -0.19 for SCP($\mathcal{X}^p$) $-$ SCP($\mathcal{X}_{\rvc}^n$) (with a p-value of $\ll 0.0001$). On  a  population-level  SCP  was  successful  in capturing  the  difference  between  normal  and  abnormal  CXR for pleural effusion.  Specifically in un-paired differences, we observe a high mean SCP values for normal subjects \ie mean SCP($\mathcal{X}^n$) = 0.69 as compared to mean SCP for abnormal patients \ie mean SCP($\mathcal{X}^p$) = 0.42. 

\end{document}